%% file: main.tex
\documentclass[10pt,journal,compsoc]{IEEEtran}
\ifCLASSOPTIONcompsoc
\usepackage[nocompress]{cite}
\else
\usepackage{cite}
\fi
\ifCLASSINFOpdf
 \usepackage[pdftex]{graphicx}
\else
\fi
\ifCLASSOPTIONcompsoc
  \usepackage[caption=false,font=footnotesize,labelfont=sf,textfont=sf]{subfig}
\else
  \usepackage[caption=false,font=footnotesize]{subfig}
\fi
\usepackage{amsmath,amsfonts,amsthm}
\usepackage{bm}
\usepackage{multirow}
\usepackage{booktabs}
\usepackage{enumitem}
\usepackage[ruled,linesnumbered,procnumbered]{algorithm2e}
\usepackage{makecell}
\usepackage{xcolor}
\usepackage{hyperref}

\hypersetup{
    colorlinks = true,
    urlcolor   = black,
    linkcolor  = black,
    citecolor  = black
}

\newtheorem{definition}{Definition}
\newcommand{\mcell}[2]{\makecell[c]{#1 \\ #2}}

\begin{document}

\newcommand{\yue}[1]{{{\textcolor{black}{\textbf{Yue:}}}{\textcolor{red}{\textbf{#1}}}}}
\newcommand{\chang}[1]{{{\textcolor{black}{\textbf{Chang:}}}{\textcolor{blue}{\textbf{#1}}}}}
\newcommand{\ping}[1]{{{\textcolor{black}{\textbf{Ping:}}}{\textcolor{green}{\textbf{#1}}}}}

\def\modelname{{MTGAN}}
\def\modelshortname{{M}}
%
\title{Multi-Label Clinical Time-Series Generation via Conditional GAN}
%
%
%
%

\author{Chang~Lu,~\IEEEmembership{Member,~IEEE,}
        Chandan~K.~Reddy,~\IEEEmembership{Senior~Member,~IEEE,}
        Ping~Wang,~\IEEEmembership{Member,~IEEE,}
        Dong~Nie,
        and~Yue~Ning,~\IEEEmembership{Member,~IEEE}%
\IEEEcompsocitemizethanks{\IEEEcompsocthanksitem C. Lu, P. Wang, and Y. Ning are with the Department of Computer Science, Stevens Institute of Technology, New Jersey,
NJ, 07310.\hfil\break
E-mail: \href{mailto:clu13@stevens.edu}{clu13@stevens.edu},
\href{mailto:pwang44@stevens.edu}{pwang44@stevens.edu},
\href{mailto:yue.ning@stevens.edu}{yue.ning@stevens.edu}
\IEEEcompsocthanksitem C.K. Reddy is with the Department of Computer Science, Virginia Tech, Arlington, VA 22203.\hfil\break
E-mail: \href{mailto:reddy@cs.vt.edu}{reddy@cs.vt.edu}
\IEEEcompsocthanksitem Dong Nie is with the Department of Computer Science, University of North Carolina at Chapel Hill.\hfil\break
E-mail: 
\href{mailto:dongnie@cs.unc.edu}{dongnie@cs.unc.edu}}}

%
%

\markboth{IEEE TRANSACTIONS ON KNOWLEDGE AND DATA ENGINEERING, VOL. X, NO. X, X 2022}%
{Lu \MakeLowercase{\textit{et al.}}: Multi-Label Clinical Time-Series Generation via Conditional GAN}
\IEEEtitleabstractindextext{%
\begin{abstract}
    In recent years, deep learning has been successfully adopted in a wide range of applications related to electronic health records (EHRs) such as representation learning and clinical event prediction. However, due to privacy constraints, limited access to EHR becomes a bottleneck for deep learning research. To mitigate these concerns, generative adversarial networks (GANs) have been successfully used for generating EHR data. However, there are still challenges in high-quality EHR generation, including generating time-series EHR data and imbalanced uncommon diseases. In this work, we propose a \textbf{M}ulti-label \textbf{T}ime-series \textbf{GAN} ({\modelname}) to generate EHR and simultaneously improve the quality of uncommon disease generation. The generator of {\modelname} uses a gated recurrent unit (GRU) with a smooth conditional matrix to generate sequences and uncommon diseases. The critic gives scores using Wasserstein distance to recognize real samples from synthetic samples by considering both data and temporal features. We also propose a training strategy to calculate temporal features for real data and stabilize GAN training. Furthermore, we design multiple statistical metrics and prediction tasks to evaluate the generated data. Experimental results demonstrate the quality of the synthetic data and the effectiveness of {\modelname} in generating realistic sequential EHR data, especially for uncommon diseases.
\end{abstract}

}

\maketitle

\IEEEdisplaynontitleabstractindextext

%
\IEEEpeerreviewmaketitle

\input{sections/introduction}
\input{sections/relatedwork}
\input{sections/method}
\input{sections/setup}
\input{sections/experiment}

\section{Conclusion}
\label{sec:conclusion}
GAN-based models are commonly adopted to generate high-quality EHR data. To tackle the challenges of generating EHR with GAN, we proposed {\modelname} to generate time-series visit records with uncommon diseases. {\modelname} can preserve temporal information as well as increase the generation quality of uncommon diseases in generated EHR by developing a temporally correlated generation process with a smooth conditional matrix. Our experimental results showed that the synthetic EHR data generated by {\modelname} not only have better statistical properties, but also achieve better results than the state-of-the-art GAN models with regards to the performance of predictive models on multiple tasks, especially for predicting uncommon diseases.

In this work, we mainly focused on GAN models to generate diseases, i.e., multi-label generation. Therefore, one of the shortcomings of {\modelname} is that it does not consider other feature types in EHR, such as procedures, medications, or lab tests. In the future, we plan to explore effective methods to generate real values including lab tests and vital signs of patients. Furthermore, we will utilize the GAN method to deal with missing values in the EHR data.


%

%

\ifCLASSOPTIONcompsoc
\section*{Acknowledgments}
This work is supported in part by the US National Science Foundation under grants 1838730, 1948432, and 2047843. Any opinions, findings, and conclusions or recommendations expressed in this material are those of the authors and do not necessarily reflect the views of the National Science Foundation.
\else
\section*{Acknowledgment}
\fi

\ifCLASSOPTIONcaptionsoff
\newpage
\fi



%
\bibliographystyle{IEEEtran}
\bibliography{ref}
    
    
    

\end{document}

%% file: sections/introduction.tex
\IEEEraisesectionheading{\section{Introduction}\label{sec:introduction}}
\IEEEPARstart{T}{he} application of electronic health records (EHR) in healthcare facilities not only automates access to key clinical information of patients, but also provides valuable data resources for researchers. To analyze EHR data, deep learning has achieved tremendous success on various tasks such as representation learning for patients and medical concepts~\cite{zhang2018patient2vec, lu2021collaborative, xu2016patienttkde}, predicting health events such as diagnoses and  mortality~\cite{darabi2020taper, choi2017gram, lu2021self, lu2021context, an2021predictiontkde}, clinical note analysis~\cite{huang2019empirical}, privacy protection~\cite{kamran2011informationtkde, ma2021datatkde}, and phenotyping~\cite{bai2018ehr, song2018attend, yin2020learningtkde}. Although EHR data are widely used in various healthcare applications, it is typically arduous for researchers to access them. On the one hand, most EHR data are not publicly available because they contain sensitive clinical information of patients, such as demographic features and diagnoses. On the other hand, some public EHR datasets including MIMIC-III~\cite{johnson2016mimic} and eICU~\cite{pollard2018eicu} only have limited samples and may not be suitable for applying large-scale deep learning-based approaches. Therefore, the limited EHR data has become on of the major bottlenecks for data-driven healthcare studies.

\begin{figure}
    \centering
    \subfloat[Visit-level]{\includegraphics[scale=0.8]{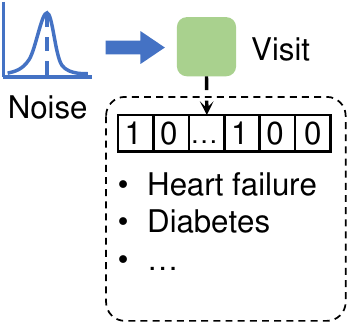}\label{fig:intro_a}} \hfill
    \subfloat[Patient-level]{\includegraphics[scale=0.8]{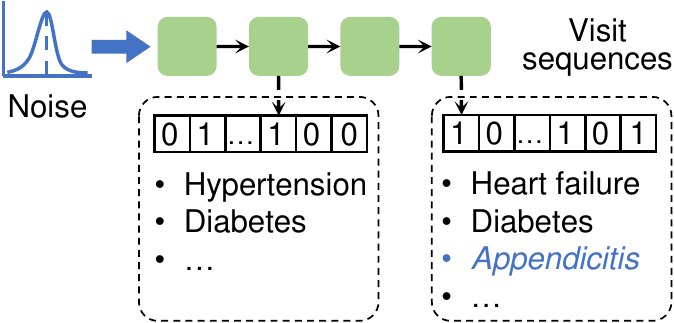}\label{fig:intro_b}}
    \caption{Visit-level generation v.s. patient-level generation. 1 or 0 in the multi-label diagnosis vector denotes the occurrence of the corresponding disease in this visit. Here, \textit{Appendicitis} is an uncommon disease.}
    \label{fig:intro}
\end{figure}

Recently, generative adversarial networks (GANs)~\cite{goodfellow2014generative} have been successful in high-quality image generation. Compared with conventional generative models such as autoencoder and variational autoencoder~\cite{kingma2013auto,lee2020generating,sun2021generating}, GANs are able to generate more realistic data~\cite{gui2021reviewtkde}. Therefore, GANs have also been applied to generate EHR~\cite{choi2017generating,baowaly2019synthesizing,zhang2020ensuring}. However, when generating EHR using existing GANs, there are still several challenges:

\begin{table*}
    \centering
    \caption{Comparison of GANs for generating EHR.}
    \label{tab:comp}
    \begin{tabular}{l|cccccc|c}
        \toprule
        \textbf{Properties}                & \mcell{\textbf{medGAN}}{\cite{choi2017generating}} & \mcell{\textbf{CTGAN}}{\cite{xu2019modeling}} & \mcell{\textbf{EMR-WGAN}}{\cite{zhang2020ensuring}} & \mcell{\textbf{RDP-CGAN}}{\cite{torfi2022differentially}} & \mcell{\textbf{TimeGAN}}{\cite{yoon2019time}} & \mcell{\textbf{T-CGAN}}{\cite{ramponi2018t}} & \mcell{\textbf{{\modelname}}}{(Proposed)}  \\ \midrule
        Time-series data generation     &$\times$& $\times$ & $\times$ & $\times$ &\checkmark&\checkmark&\checkmark\\
        Preserving temporal correlations   &$\times$& $\times$ & $\times$ & $\times$ &\checkmark& $\times$ &\checkmark\\
        Uncommon diseases generation        &$\times$&\checkmark& $\times$ & $\times$ & $\times$ & $\times$ &\checkmark\\
        Stable training with sparse EHR &$\times$& $\times$ & $\times$ & $\times$ & $\times$ & $\times$ &\checkmark\\
        \bottomrule
    \end{tabular}
\end{table*}

\begin{enumerate}[wide]
\item \textbf{Generating time-series EHR data}. In EHR data, a patient can have multiple visits. However, most existing GANs for generating EHR, such as medGAN~\cite{choi2017generating}, EMR-WGAN~\cite{zhang2020ensuring}, Smooth-GAN~\cite{rashidian2020smooth}, and RDP-CGAN~\cite{torfi2022differentially}, can only generate independent visits instead of time-series data. It is because traditional GANs designed for image generation only generate one image given a noise input. Although it is possible to combine generated visits randomly as a sequence, this method cannot preserve temporal information to disclose patient-level features. \figurename~\ref{fig:intro} shows an example of visit-level and patient-level data generation.  In \figurename~\ref{fig:intro}\subref{fig:intro_a}, it generates diagnoses for only one visit. An ideal sequence generation is described in \figurename~\ref{fig:intro}\subref{fig:intro_b}. The diagnoses in two close visits are similar and related, such as hypertension and heart failure. Recently, SeqGAN~\cite{yu2017seqgan} and TimeGAN~\cite{yoon2019time} are proposed to generate sequences as a simulation of sentences. However, unlike words in a sentence, each time step (visit) of EHR contains multi-label variates (i.e., diagnoses shown in \figurename~\ref{fig:intro}). Therefore, generating multi-label time-series EHR with temporal correlations still remains a challenge.

\item \textbf{Generating uncommon diseases}. Based on the statistics of a well-known public EHR dataset, MIMIC-III, some diseases are frequently diagnosed, such as hypertension and diabetes, while some other diseases such as tuberculosis are less common. Although these diseases do not frequently occur, it is still valuable to study them to provide better care plans for patients,~e.g., analyzing occurrence patterns to improve diagnosis prediction accuracy. Despite the ability of existing GANs to generate time-series EHR data, it is still challenging for them to learn a good distribution for uncommon diseases. Instead of only generating frequent diseases shown in \figurename~\ref{fig:intro}\subref{fig:intro_a}, we need to find effective ways to generate uncommon diseases, such as \textit{Appendicitis} in \figurename~\ref{fig:intro}\subref{fig:intro_b} given highly imbalanced EHR datasets.

\item \textbf{Evaluating synthetic EHR data}. Since EHR datasets have an imbalanced disease distribution, traditional evaluation metrics for synthetic images such as Kullback-Leibler divergence and Jensen-Shannon divergence do not provide sufficient attention to uncommon diseases. As a result, we may still get low divergence between the distribution of real and synthetic EHR data when they are close in terms of diseases with higher frequency. Therefore, it is still necessary to explore appropriate metrics to evaluate the quality of synthetic EHR data, especially for uncommon diseases.
\end{enumerate}

To address these challenges, we propose {\modelname}, a \textbf{m}ulti-label \textbf{t}ime-series generation model using a conditional GAN to simultaneously generate time-series diagnoses and uncommon diseases. In the generator, we first propose to recursively generate patient-level diagnosis probabilities with a gated recurrent unit (GRU). Then, to generate uncommon diseases, we adopt the idea of the conditional vector in CTGAN~\cite{xu2019modeling} and broadcast this vector into a smooth conditional matrix throughout all visits in sequences. In the critic of {\modelname}, we propose to discriminate real and synthetic samples by giving scores to both the data and their temporal features. Finally, we design a training strategy to optimize {\modelname} by sampling discrete diseases from visit-level probabilities and forming the patient-level visit sequences to stabilize the training process. The model computes temporal features of real data by pre-training a GRU with the task of next visit prediction. The contributions of this work are summarized as follows:
\begin{itemize}[leftmargin=*]
	\item We propose a time-series generative adversarial network {\modelname} to generate multi-label patient-level EHR data.
	The generator, critic, and training strategy of {\modelname} are able to simultaneously generate realistic visits and preserve temporal correlations across different visits.
	\item We propose a smooth conditional matrix to cope with the imbalanced disease distribution in EHR data and improve the generation quality of uncommon diseases.
	\item We use multiple statistical metrics for synthetic EHR evaluation and design a normalized distance especially for uncommon diseases. Meanwhile, we verify that the synthetic EHR generated by {\modelname} can boost deep learning models on temporal health event prediction tasks.
\end{itemize}

The remaining parts of this paper are listed below: We first discuss related work about EHR generation in Section~\ref{sec:related_work}. Then, we formulate the EHR generation problem in Section~\ref{sec:problem} and introduce the details of {\modelname} in Section~\ref{sec:method}. Next, the experimental setups and results are demonstrated in Sections~\ref{sec:exp_setup} and \ref{sec:exp}, respectively. Finally, we summarize this paper and discuss the future work in Section~\ref{sec:conclusion}.

%% file: sections/relatedwork.tex
\section{Related Work}
\label{sec:related_work}
\subsection{Generative Adversarial Networks}
The generative adversarial networks are first proposed by Goodfellow \textit{et al.}~\cite{goodfellow2014generative} to generate realistic images. A typical GAN contains a generator to generate synthetic samples and a discriminator to distinguish real samples from generated samples. Arjovsky \textit{et al.}~\cite{arjovsky2017wasserstein} propose WGAN by replacing the binary classification in the discriminator with the Wasserstein distance to alleviate mode collapse and vanishing gradient in GAN. Gulrajani \textit{et al.}~\cite{gulrajani2017improved} introduce a gradient penalty in WGAN-GP to improve the training of WGAN. Xu \textit{et al.}~\cite{xu2019modeling} propose CTGAN to generate imbalanced tabular data with a conditional vector. Wang \textit{et al.}~\cite{wang2019learningtkde} propose a graph softmax method in GraphGAN to sample discrete graph data. Unfortunately, typical GANs are not able to generate time-series data, and therefore cannot be directly applied to generate EHR data.

\subsection{GANs for Sequence Generation}
To generate sequences with discrete variates, SeqGAN~\cite{yu2017seqgan} is proposed by Yu \textit{et al.} with the REINFORCE algorithm and policy gradient. Yoon \textit{et al.}~\cite{yoon2019time} propose TimeGAN by jointly training with a GAN loss, a reconstruction loss, and a sequential prediction loss. To generate time-series data with conditions, Ramponi \textit{et al.}~\cite{ramponi2018t} propose T-CGAN by specifying the time step of a data sample as the condition. Esteban \textit{et al.}~\cite{esteban2017real} propose a recurrent conditional GAN, RCGAN, to generate real value medical data. Du \textit{et al.}~\cite{du2021gantkde} propose a GAN-based anomaly detection algorithm for multivariate time series data. Liu \textit{et al.}~\cite{liu2022timetkde} also apply the GAN framework in BeatGAN by adding an encoder and decoder to reconstruct time-series data for anomaly detection. However, generating multi-label synthetic data from imbalanced datasets is not considered in SeqGAN and TimeGAN. For T-CGAN, when generating a sample, it only uses the temporal position of this sample as the condition, thus ignoring temporal correlations of the entire sequence and the imbalanced distribution of labels. For RCGAN and BeatGAN, they are designed for real-value time-series data and do not fit for EHR data.

\subsection{Generating EHR with GANs}
To generate sequential EHR data, Lee \textit{et al.}~\cite{lee2020generating} and Sun \textit{et al.}~\cite{sun2021generating} leverage adversarial autoencoder~\cite{makhzani2015adversarial} with a sequence-to-sequence autoencoder. However, compared to autoencoders, GANs allow for more flexibility and diversity in generating samples. Gong \textit{et al.}~\cite{gong2022diffuseq} propose DiffSeq to generate text sequences based on the diffusion model~\cite{ho2020denoising}. Unfortunately, training diffusion models requires large-scale datasets to achieve stable training~\cite{moon2022fine}, which may not be suitable for EHR generation.

Recently, GANs are applied to generate EHR to address the problem of limited data sources in healthcare applications. Che \textit{et al.}~\cite{che2017boosting} propose ehrGAN by feeding the generator with masked real data to generate EHR data. Choi \textit{et al.}~\cite{choi2017generating} propose medGAN by introducing an auto-encoder. The generator outputs a latent feature, and medGAN uses the auto-encoder to decode synthetic data from the latent feature. Baowaly \textit{et al.}~\cite{baowaly2019synthesizing} replace the GAN framework in medGAN with WGAN-GP and propose medWGAN. EMR-WGAN is proposed by Zhang \textit{et al.}~\cite{zhang2020ensuring}. It removes the auto-encoder in medGAN and let the generator directly output synthetic data. Torfi \textit{et al.}~\cite{torfi2022differentially} propose RDP-CGAN with a convolutional auto-encoder and convolutional GAN. The RDP-CGAN model also uses a differential privacy method to preserve privacy in the synthetic EHR data.

However, these GANs only generate single visits instead of patient-level data. As a result, the synthetic EHR data generated by these GANs cannot be used for many time-series tasks such as temporal health event prediction. In addition, they do not consider uncommon diseases given that the diseases in EHR datasets are usually imbalanced, which decreases the quality of the synthetic EHR data. Furthermore, a majority of GANs for EHR generation are not stable when dealing with sparse EHR data, which may make it difficult to train the GANs. In general, we compare related GANs in Table~\ref{tab:comp} based on the important properties required for generating EHR. In this work, we simultaneously consider generating patient-level EHR data and uncommon diseases.

%% file: sections/method.tex
\begin{table}[]
    \centering
    \caption{Notations used in this paper.}
    \label{tab:symbols}
    \begin{tabular}{l|l}
        \toprule
        \textbf{Notation} & \textbf{Explanation} \\
        \midrule
        $\mathbf{x}_t$ & Diagnosis vector of the $t$-th visit \\
        $G, D$ & Generator and Critic \\
        $\mathcal{D}, \tilde{\mathcal{D}}$ & Real and generated EHR datasets \\
        $\boldsymbol{x}, \tilde{\boldsymbol{x}}$ & Real and generated sample \\
        $\tilde{\mathbf{P}}$ & Generated probability distribution for diseases \\
        $\mathbf{H}, \tilde{\mathbf{H}}$ & Real and generated hidden states \\
        $\mathbf{c}$ & Smooth conditional matrix \\
        $g_{\text{gru}}$ & The GRU model in the generator \\
        $g'_{\text{gru}}$ & Pre-trained GRU model for real EHR data \\
        \bottomrule
    \end{tabular}
\end{table}

\begin{figure}
    \centering
    \includegraphics[width=\linewidth]{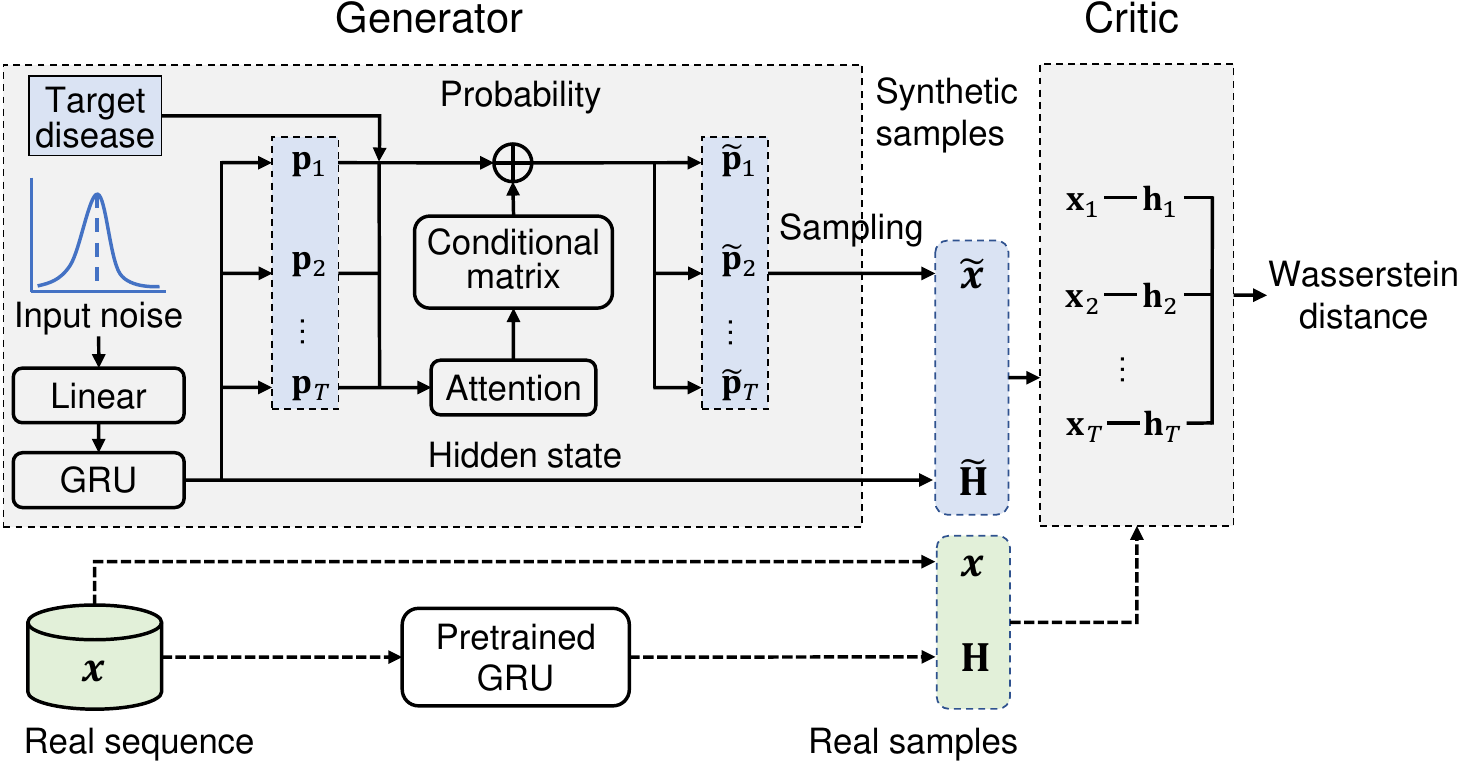}
    \caption{The model overview of {\modelname}. The generator uses a GRU to obtain patient-level diagnosis probabilities with hidden states and use the probabilities to calculate attention scores as a conditional matrix to generate the target disease. The critic calculates a Wasserstein distance by considering both synthetic/real samples and their temporal features.}
    \label{fig:system}
\end{figure}

\section{Problem Formulation}
\label{sec:problem}
In this section, we describe the EHR dataset in detail and formally define the research problem, EHR generation. In addition, we list important symbols and their corresponding explanations in Table~\ref{tab:symbols}.

An EHR dataset consists of visit sequences of patients to healthcare facilities. A visit contains different data types, such as diagnoses, procedures, lab tests, and clinical notes. An important feature in EHR data is diagnoses represented by disease codes, such as ICD-9~\cite{icd9cm} or ICD-10~\cite{icd10}. In this work, we focus on generating diagnoses, and the research questions is formulated into a time-series multi-label generation problem. To describe an EHR dataset, we first give the following definitions:

\begin{definition}[Visit]
A visit contains one or multiple diagnoses. The $t$-th visit is denoted by a binary vector $\mathbf{x}_t \in \{0, 1\}^d$, where $d$ is the number of distinct diseases, i.e., disease types in the EHR dataset. $\mathbf{x}_t^i = 1$ means the disease $i$ is diagnosed in the $t$-th visit.
\end{definition}

\begin{definition}[Visit sequence]
Given a patient $u$, the visit sequence of this patient is denoted as $\boldsymbol{x}_u = (\mathbf{x}_1, \mathbf{x}_2, \dots, \mathbf{x}_T) \in \{0, 1\}^{d \times T}$, where $T$ is the sequence length.
\end{definition}

\begin{definition}[EHR dataset]
An EHR dataset $\mathcal{D}$ is a collection of visit sequences: $\mathcal{D} = \{\boldsymbol{x}_u \mid u \in \mathcal{U}\}$, where $\mathcal{U}$ is a patient set.
\end{definition}

Based on these descriptions of EHR data, the EHR generation problem is defined as below:

\begin{definition}[Problem formulation]
Given a real EHR dataset $\mathcal{D}$, we aim to generate a synthetic EHR dataset $\tilde{\mathcal{D}}$ such that $\tilde{\mathcal{D}}$ has the following properties:
\begin{enumerate}
    \item The disease distribution of $\tilde{\mathcal{D}}$ is close to $\mathcal{D}$.
    \item The disease type in $\tilde{\mathcal{D}}$ is similar to $\mathcal{D}$ when $|\mathcal{D}| = |\tilde{\mathcal{D}}|$.
\end{enumerate}
Here, $|\cdot|$ denotes the number of data samples.
\end{definition}

\section{The Proposed \modelname~Model}
\label{sec:method}
In this section, we introduce some preliminaries about GANs and discuss the proposed {\modelname} to generate discrete diagnoses in electronic health records, including detailed challenges in generating EHR data and our proposed generator, critic, and the training strategy.
The model overview of {\modelname} is shown in~\figurename{~\ref{fig:system}}.

\subsection{Preliminaries of Generative Adversarial Networks}
In a typical framework of generative adversarial networks (GANs), there exists a generator $G$ that takes a noise $\boldsymbol{z} \in \mathbb{R}^{s}$ from a random distribution as the input and generates a synthetic data sample $\tilde{\boldsymbol{x}} = G(\boldsymbol{z})$. The discriminator $D$ is another key part of GANs. It tries to distinguish real data samples $\boldsymbol{x}$ from generated samples $\tilde{\boldsymbol{x}}$. The underlying mechanism of GANs can be formulated as a min-max game: the generator tries to generate realistic samples to deceive the discriminator and let it think $\tilde{\boldsymbol{x}}$ is real; the discriminator conducts a binary classification and tries to classify all real and synthetic samples correctly. A vanilla GAN is optimized using the following loss function:
\begin{align}
    \underset{G}{\min}~\underset{D}{\max}~\underset{{\boldsymbol{x} \sim p_{\boldsymbol{x}}}}{\mathbb{E}}[\log{D(\boldsymbol{x})}] + \underset{{\boldsymbol{z} \sim p_{\boldsymbol{z}}}}{\mathbb{E}}[\log{(1 - D(G(\boldsymbol{z})))}].
\end{align}

However, such a simple GAN is sometimes hard to train due to the vanishing gradient problem, mode collapse, and failure to converge. To address these issues, Arjovsky \textit{et al.}~\cite{arjovsky2017wasserstein} use the Wasserstein distance in WGAN to train the generator and discriminator (called a critic in WGAN). Gulrajani \textit{et al.}~\cite{gulrajani2017improved} introduce a gradient penalty for training the critic in WGAN-GP. The updated loss functions to train the generator and critic respectively are as follows:
\begin{align}
    L_D =&~\underset{{\boldsymbol{z} \sim p_{\boldsymbol{z}}}}{\mathbb{E}}[D(G(\boldsymbol{z}))] - \underset{{\boldsymbol{x} \sim p_{\boldsymbol{x}}}}{\mathbb{E}}[D(\boldsymbol{x})] \notag \\
    &~+ \lambda \underset{{\hat{\boldsymbol{x}} \sim p_{\hat{\boldsymbol{x}}}}}{\mathbb{E}}\left[\left(\|\Delta_{\hat{\boldsymbol{x}}}D(\hat{\boldsymbol{x}})\|_2 - 1\right)^2\right], \\
    L_G =&~-\underset{{\boldsymbol{z} \sim p_{\boldsymbol{z}}}}{\mathbb{E}}[D(G(\boldsymbol{z}))],
\end{align}
where $L_G$ and $L_D$ are the losses for the generator and critic, respectively; $\lambda$ is a coefficient for the gradient penalty; $\hat{\boldsymbol{x}} = \epsilon \boldsymbol{x} + (1 - \epsilon)\tilde{\boldsymbol{x}}$, $\epsilon \sim \text{U}[0, 1]$ is sampled from a uniform distribution; $\Delta$ denotes the derivation operation; and $\|\cdot\|_2$ means $\ell^2$-norm. In these two loss functions, $D(\cdot)$ calculates a critic score for an input. It tries to maximize the score for real data and minimize the score for synthetic data. It turns the binary classification of the original GAN into a regression problem. By introducing Wasserstein distance and gradient penalty, training GAN can be more stable. Therefore, similar to EMR-WGAN~\cite{zhang2020ensuring} and Smooth-GAN~\cite{rashidian2020smooth}, we also introduce the gradient penalty in the training of WGAN.

\subsection{Generator}
\label{sec:generator}
As we discussed before, to generate realistic EHR samples, we must address the following specific challenges:
\begin{enumerate}[label=\textbf{C\arabic*:}, ref=\textbf{C\arabic*}]
    \item How to incorporate temporal features of visit sequences to increase the correlation of adjacent visits?\label{challenge_gc1}
    \item How to generate uncommon diseases in the real EHR dataset $\mathcal{D}$ with an unbiased distribution?\label{challenge_gc2}
\end{enumerate}

\subsubsection{Temporally-Correlated Probability Generation}
In TimeGAN~\cite{yoon2019time}, when generating sequences, an intuitive method is using recurrent neural networks (RNN). In each time step, the input of the RNN cell is a random noise and the hidden state passed from the previous time step. The output of each time step is a new hidden state. We can use the hidden state to generate each visit and combine all visits as a sequence. However, we think that using noises to generate visits for every time step may somewhat bring uncontrollable randomness and weaken the temporal correlation between adjacent visits. We believe an optimized generator is able to generate the entire sequence given a single noise vector at the beginning of the sequence. Similar to the temporal health event prediction task studied in GRAM~\cite{choi2017gram}, CGL~\cite{lu2021collaborative}, and Chet~\cite{lu2021context}, a good generator should predict (generate) the diagnoses in the next visit, given all previous visits. Therefore, based on this idea, we propose to recursively generate the visit sequence from a single noise vector $\boldsymbol{z}$, in order to increase the temporal correlation of adjacent visits, i.e., the challenge \ref{challenge_gc1}.

Given a random noise vector $\boldsymbol{z} \in \mathbb{R}^s$ and a visit length $T$, since the disease values in each visit is 0 or 1, we first generate the disease probability $\mathbf{P}_1$ in the first visit by decoding the noise vector:
\begin{align}
    \mathbf{P}_1 = \sigma(\mathbf{W}\boldsymbol{z}) \in \mathbb{R}^d.
\end{align}
Here, $\mathbf{W} \in \mathbb{R}^{d \times s}$ is the weight to project the noise into the visit space. $\sigma$ is the sigmoid function. After having the first visit, we can recursively generate the disease probability of remaining visits using a gated recurrent unit (GRU)~\cite{chung2014empirical} $g_{\text{gru}}$:
\begin{align}
    \tilde{\mathbf{h}}_{t} &= g_{\text{gru}}(\mathbf{P}_t, \tilde{\mathbf{h}}_{t - 1}) \in \mathbb{R}^s, \\
    \mathbf{P}_{t + 1} &= \sigma(\mathbf{W}\tilde{\mathbf{h}}_{t}) \in \mathbb{R}^{d}.
\end{align}
Here $\tilde{\mathbf{h}}_t$ denotes the hidden state of GRU at the time step $t$. We set $\tilde{\mathbf{h}}_0 = \mathbf{0}$ and set the noise dimension to be the same as hidden units of GRU, because we regard the noise vector as the initial hidden state. Next, we use GRU to calculate the hidden state of the time step $t$ using the hidden state of $t - 1$ and the generated visit probability $\mathbf{P}_t$. Then, we use the same decoding for $\boldsymbol{z}$ to generate $\mathbf{P}_{t + 1}$ for the visit $t + 1$. Finally, we combine all the generated disease probabilities as a patient-level distribution $\mathbf{P}$ for a synthetic EHR data sample: $\mathbf{P} = (\mathbf{P}_1, \mathbf{P}_2, \dots, \mathbf{P}_T) \in \mathbb{R}^{d \times T}$.

\subsubsection{Smooth Conditional Matrix}

After generating the patient-level probabilities, we need to address the challenge of generating uncommon diseases, i.e., \ref{challenge_gc2}. To deal with highly imbalanced tabular data, CTGAN~\cite{xu2019modeling} is proposed to use conditional vectors to guide the GAN training process. 
More specifically, it first specifies a category for a  tabular feature as the target category. Then, it uses a conditional vector where the corresponding entry for the target category is 1. Finally, it concatenates the conditional vector with the noise vector as the generator input to generate samples that belong to the target category.

Inspired by CTGAN, we aim to specify a target disease and adopt the conditional vector to generate a visit sequence that contains the target disease. However, CTGAN is designed for non-sequential data. For a visit sequence, the target disease may appear in one or multiple visits. If we directly concatenate the conditional vector with the noise vector, this input will have the highest impact on the first visit and a decreasing impact on the remaining visits, due to the characters of RNN-based models. As a result, it is highly possible that this disease only appears in the first visit. If we concatenate the conditional vector for all $\mathbf{P}_t$, the generator may output a visit sequence where each visit contains the target disease. To avoid these extreme cases, we propose to smooth the conditional vector into a conditional matrix $\mathbf{c} \in \mathbb{R}^{d \times T}$ for all visits. First, we apply a location-based attention method~\cite{luong2015effective} to the generated probability to broadcast the target disease $i$ into a probability distribution (attention score) for all visits:
\begin{align}
    v_t = \mathbf{W}_v\mathbf{P}_t \in \mathbb{R}, ~\text{where}~t \in \{1, 2, \dots, T\},
\end{align}
\begin{align}
    \text{score}_t = \frac{e^{v_t}}{\sum_{\tau=1}^{T}{e^{v_{\tau}}}}.
\end{align}
Here, $\mathbf{W}_v \in \mathbb{R}^{1 \times d}$ is an attention weight, and $\sum_{t=1}^{T}{\text{score}_t} = 1$. With this score, if the generator assigns a higher probability of the target disease to the visit $t$, the GAN model can generate corresponding co-occurred diseases in this visit. After calculating the score for each visit, we create a conditional matrix $\mathbf{c} \in \mathbf{0}^{d \times T}$, and set the entry $\mathbf{c}_{i, t}$ corresponding to the target disease $i$ and visit $t$ as $\text{score}_t$:
\begin{align}
    \mathbf{c}_{i, t} = \text{score}_t.
\end{align}
Then, we use $\mathbf{c}$ to calibrate the generated probability by adding $\mathbf{c}$ to $\mathbf{P}$ and get a calibrated probability $\tilde{\mathbf{P}}$:
\begin{align}
    \tilde{\mathbf{P}} = \min{(1, \mathbf{P} \oplus \mathbf{c})} \in \mathbb{R}^{d \times T}.
\end{align}
Here, $\oplus$ denotes an element-wise sum of two matrices. We also clip $\tilde{\mathbf{P}}$ to make sure it is no greater than 1. In this way, the target disease is smoothed to all $T$ visits. Therefore, the conditional matrix can increase the probability of target diseases and let the uncommon diseases gain more exposure.

In summary, given a noise vector $\boldsymbol{z}$ and a target disease $i$, the generator $G$ is able to generate a calibrated probability distribution $\tilde{\mathbf{P}}$ for diseases of a visit sequence: $\tilde{\mathbf{P}} = (\tilde{\mathbf{P}}_1, \tilde{\mathbf{P}}_2, \dots, \tilde{\mathbf{P}}_T) = G(\boldsymbol{z}, i)$. We will discuss how to generate discrete diagnoses in Section~\ref{sec:training}.

\subsection{Critic}
For the critic distinguishing real and fake EHR data, there is still a specific challenge to be addressed to improve the quality of synthetic samples:
\begin{enumerate}[label=\textbf{C\arabic*:}, ref=\textbf{C\arabic*}, resume]
    \item How to calculate a sequential Wasserstein distance for real and synthetic visit sequences? \label{challenge_dc1}
\end{enumerate}

Given a visit sequence, an optimized critic should consider two aspects to determine whether this sequence is real or not. The first is whether each visit in a sequence is real. The second is whether this visit sequence is able to reflect temporally-correlated characters. These two aspects are intuitive because a visit sequence looks real only if each independent visit looks real. Furthermore, even if each visit looks real, the entire sequence may not be real. For example, we exchange two visits from two different patients or from the same patient. Even though each visit is real, the critic should still detect the abnormal visit sequence if the two exchanged visits are largely different.

Based on the above analysis, we propose a \textbf{\textit{sequential critique}} that can simultaneously distinguish whether individual visits are real and whether the entire sequence are real. Given an input sequence $\boldsymbol{x} = (\mathbf{x}_1, \mathbf{x}_2, \dots, \mathbf{x}_T) \in \{0, 1\}^{d \times T}$ and the temporal features of this sequence $\mathbf{H} = (\mathbf{h}_1, \mathbf{h}_2, \dots, \mathbf{h}_T) \in \mathbb{R}^{d \times T}$ that correspond to each visit, the critic first concatenates the diagnosis vector $\mathbf{x}_t$ and temporal feature vector $\mathbf{h}_t$ for each visit. Then it uses a multi-layer perceptron (MLP) to calculate a critic score for this visit. Finally, the score $r$ for the sequence is an average of all visits. This process can be summarized as follows:
\begin{align}
    \mathbf{m}_t = \mathbf{x}_t~||~\mathbf{h}_t \in \mathbb{R}^{d + s}, \label{eq:critic_cat}
\end{align}
\begin{align}
    r = \frac{1}{T}{\sum_{t = 1}^{T}{\text{MLP}(\mathbf{m}_t)}} \in \mathbb{R}.
\end{align}
Here, $||$ denotes the concatenation operation. In this equation, We use the average of all visits because we hypothesize $\mathbf{m}_t$ contains the temporal feature of each visit and therefore is capable of distinguishing time-series data. In this way, the critic can simultaneously consider individual visits and the temporal correlation of adjacent visits. Note that, the visit sequence $\boldsymbol{x}$ can be either a real or a generated sequence.

In summary, given an input sequence $\boldsymbol{x}$ and the temporal features $\mathbf{H}$ of $\boldsymbol{x}$, the critic $D$ computes a score for this sequence: $r = D(\boldsymbol{x}, \mathbf{H})$.

\subsection{Training Strategy}
\label{sec:training}
After defining the generator and the critic, there are still two remaining problems when generating diseases and training the critic with real/synthetic samples and temporal features:
\begin{enumerate}[leftmargin=*]
    \item How to get temporal features of real samples?
    \item How to obtain discrete diagnoses from the generated probability distribution?
\end{enumerate}

\subsubsection{Temporal Feature Pre-training }
For the first problem, when generating the probability distribution, we have already got the hidden state $\tilde{\mathbf{h}}_t$ for each generated visit. We conjecture that if the generator is optimized, the distribution of hidden state for generated visits should also be consistent with real samples. Therefore, we design a prediction task to pre-train a base GRU to calculate the hidden state for real samples. Given a real visit sequence $\boldsymbol{x} = (\mathbf{x}_1, \mathbf{x}_2, \dots, \mathbf{x}_T) \in \{0, 1\}^{d \times T}$, we aim to use a GRU $g'_{\text{gru}}$ that has an identical structure to $g_{\text{gru}}$ to predict the next visit for each $\mathbf{x}_t$ in $\boldsymbol{x}$. To do this, we first transform $\boldsymbol{x}$ into a feature sequence $(\mathbf{x}_1, \mathbf{x}_2, \dots, \mathbf{x}_{T - 1}) \in \{0, 1\}^{d \times (T - 1)}$ and a label sequence $(\mathbf{y}_1, \mathbf{y}_2, \dots, \mathbf{y}_{T - 1}) \in \{0, 1\}^{d \times (T - 1)}$, where $\mathbf{y}_t = \mathbf{x}_{t + 1}$. We then use the $g'_{\text{gru}}$ to calculate the hidden state $\mathbf{h}_t$ for $\mathbf{x}_t$ and predict the next visit $\hat{\mathbf{y}}_t$:
\begin{align}
    \mathbf{h}_t &= g'_{\text{gru}}(\mathbf{x}_t, \mathbf{h}_{t - 1}) \in \mathbb{R}^s, \\
    \hat{\mathbf{y}}_{t} &= \sigma(\mathbf{W}'\mathbf{h}_{t}) \in \mathbb{R}^{d}.
\end{align}
Here, we also set $\mathbf{h}_0 = \mathbf{0}$. To pre-train the $g'_{\text{gru}}$, we use a binary cross-entropy loss for a single visit prediction, and calculate the sum of all visits as the final loss $L_{\text{pre}}$:
\begin{align}
    L_{\text{pre}} = \sum_{t = 1}^{T}{\sum_{i = 1}^{d}{\mathbf{y}_i\log{\hat{\mathbf{y}}_i}} + (1 - \mathbf{y}_i)\log{(1 - \hat{\mathbf{y}}_i)}}
\end{align}

After getting the pre-trained $g'_{\text{gru}}$, we freeze its parameters and use it to calculate the temporal features $\mathbf{H} = (\mathbf{h}_1, \mathbf{h}_2, \dots, \mathbf{h}_T) = g'_{\text{gru}}(\boldsymbol{x}) \in \mathbb{R}^{d \times T}$ for real samples in the critic. For the synthetic data, we let the generator $G$ return both the probability $\tilde{\mathbf{P}}$ and the hidden state $\tilde{\mathbf{H}} = (\tilde{\mathbf{h}}_1, \tilde{\mathbf{h}}_2, \dots, \tilde{\mathbf{h}}_T) \in \mathbb{R}^{d \times T}$: $(\tilde{\mathbf{P}}, \tilde{\mathbf{H}}) = G(\boldsymbol{z}, i)$.

\subsubsection{Discrete Disease Sampling}
In Section~\ref{sec:generator}, our generator outputs a probability distribution of visit sequences. When training the generator, it is reasonable to directly feed the probability distribution to the critic because we aim to let the generator increase the probability of occurred diseases and decrease the probability of unoccurred diseases based on gradients flowed from the critic. However, if we directly use the probability distribution to train the critic, it will increase the uncertainty of the generator and make the training less stable. For example, let us say that the generator gives a probability of 0.8 for a disease. After training the critic, it gives a lower score for this generation. When training the generator in the next step, it only knows 0.8 leads to a lower score but does not know whether to increase or decrease the probability to reach a higher score. As a consequence, we may need many iterations to make the training of generator converge after a lot of explorations in the input space. To stabilize the training process, we propose to train the critic by sampling from the generated distribution $\tilde{\mathbf{P}}$ to get a discrete diagnoses sequence $\tilde{\boldsymbol{x}} = (\tilde{\mathbf{x}}_1, \tilde{\mathbf{x}}_2, \dots, \tilde{\mathbf{x}}_T) \sim \tilde{\mathbf{P}} \in \{0, 1\}^{d \times T}$, where
\begin{align}
    \tilde{\mathbf{x}}_t \boldsymbol{\sim} \text{Bernoulli}(\tilde{\mathbf{P}}_t) \in \{0, 1\}^{d}.
\end{align}
Here, we use $\boldsymbol{\sim}$ to denote element-wise sampling, and $\text{Bernoulli}(p)$ means sampling from a Bernoulli distribution with the success probability as $p$. In this approach, the synthetic data for training the critic are discrete. We also use the probability 0.8 as an example. Assume the sampled output is 1, after a generator optimization step, it not only knows 0.8 will get a low score, but also learns that it should decrease the probability to reach a higher score.

There is another advantage of generating discrete diseases by sampling. In traditional GANs for generating EHR such as medGAN~\cite{choi2017generating}, medWGAN~\cite{baowaly2019synthesizing}, Smooth-GAN~\cite{rashidian2020smooth}, and RDP-CGAN~\cite{torfi2022differentially}, after getting the disease probability from either the generator or autoencoder, they directly round the probability to get the discrete diseases. However, for uncommon diseases, the probabilities of them are usually low. Rounding the probability will further decrease the frequency of uncommon diseases in generated samples. Therefore, we use sampling from the probability as another measure to generate uncommon diseases, i.e., \ref{challenge_gc2}.

Finally, we use the losses $L_G$ and $L_D$ to train the generator and critic respectively, given a target disease $i \sim \text{U}[0, d]$:
\begin{align}
    L_D =&~\mathbb{E}_{\tilde{\boldsymbol{x}} \sim \tilde{\mathbf{P}}}[D(\tilde{\boldsymbol{x}}, \tilde{\mathbf{H}}))] - \mathbb{E}_{{\boldsymbol{x} \sim p_{\boldsymbol{x}|i}}}[D(\boldsymbol{x}, \mathbf{H})] \notag \\
    &~+ \lambda{\mathbb{E}_{\hat{\boldsymbol{x}} \sim p_{\hat{\boldsymbol{x}}}, \hat{\mathbf{H}} \sim p_{\hat{\mathbf{H}}}}}\left[(\|\Delta_{\hat{\boldsymbol{x}}, \hat{\mathbf{H}}}D(\hat{\boldsymbol{x}}, \hat{\mathbf{H}})\|_2 - 1)^2\right], \label{eq:loss_d}\\
    L_G =&~-\mathbb{E}_{\boldsymbol{z} \sim p_{\boldsymbol{z}}}[D(\tilde{\mathbf{P}}, \tilde{\mathbf{H}})]. \label{eq:loss_g}
\end{align}
The pseudo-code for training {\modelname} is summarized in Algorithm~\ref{algo:model_algo}. In each iteration, we first sample a target disease $i$ from a discrete uniform distribution $\text{U}[0, d]$. When training critic at lines 3-11, we sample real data $\boldsymbol{x} \sim p_{\boldsymbol{x}|i}$ that contain this target disease in any visit, following the setting in CTGAN. When calculating the gradient penalty, besides letting $\hat{\boldsymbol{x}} = \epsilon \boldsymbol{x} + (1 - \epsilon)\tilde{\boldsymbol{x}}$, we also incorporate $\hat{\mathbf{H}} = \epsilon \mathbf{H} + (1 - \epsilon)\tilde{\mathbf{H}}$ with the same $\epsilon$ into the calculation. At lines 13-15, we train the generator by feeding the synthetic probabilities into critic. Finally, we repeat the training of the critic and the generator until they converge.

\begin{algorithm}
    \caption{\modelname-Training ($\mathcal{D}, g'_{\text{gru}}, n_{\text{critic}}$)}
    \label{algo:model_algo}
    \DontPrintSemicolon
    \SetAlgoLined
    \newcommand\mycommfont[1]{\footnotesize\ttfamily\textcolor{gray}{#1}}
    \SetCommentSty{mycommfont}
    \SetKwInOut{Input}{Input}
    \SetKwInOut{Output}{Output}
    \Input{Real EHR dataset $\mathcal{D}$\\Pre-trained GRU $g'_{\text{gru}}$\\Critic training number $n_{\text{critic}}$}
    $d \leftarrow$ Count the disease number in $\mathcal{D}$ \\
    \Repeat{convergence}{
        Sample a target disease $i \sim \text{U}[0, d]$ \\
        \tcp{Training the critic}
        \For{$j \leftarrow 1$ to $n_{\text{critic}}$}{
            Sample real data $\boldsymbol{x} \sim p_{\boldsymbol{x}|i}$, noise $\boldsymbol{z} \sim p_{\boldsymbol{z}}$, coefficient $\epsilon \sim \text{U}[0, 1]$ \\
            $\mathbf{H} \leftarrow g'_\text{gru}(\boldsymbol{x})$ \\
            $\tilde{\mathbf{P}}, \tilde{\mathbf{H}} \leftarrow G(\boldsymbol{z}, i)$ \\
            Sample discrete diseases $\tilde{\boldsymbol{x}} \sim \tilde{\mathbf{P}}$ \\
            $\hat{\boldsymbol{x}} \leftarrow \epsilon \boldsymbol{x} + (1 - \epsilon)\tilde{\boldsymbol{x}}$ \\
            $\hat{\mathbf{H}} \leftarrow \epsilon \mathbf{H} + (1 - \epsilon)\tilde{\mathbf{H}}$ \\
            Optimize the critic $D$ using $L_D$
        }
        \tcp{Training the generator}
        Sample noise $\boldsymbol{z} \sim p_{\boldsymbol{z}}$ \\
        $\tilde{\mathbf{P}}, \tilde{\mathbf{H}} \leftarrow G(\boldsymbol{z}, i)$ \\
        Optimize the generator $G$ using $L_G$ \\
    }
\end{algorithm}

%% file: sections/setup.tex
\section{Experimental Setups}
\label{sec:exp_setup}
\subsection{Evaluation Metrics}
\label{sec:metrics}
To evaluate the statistical quality of the generated EHR dataset $\tilde{\mathcal{D}}$, we use the following metrics:

\begin{itemize}[leftmargin=*, labelwidth=0.5\labelwidth]
\item \textit{Generated disease types (GT)}: We use the generated disease types to evaluate whether the GAN model can generate all diseases in $\mathcal{D}$. When $|\mathcal{D}| = |\tilde{\mathcal{D}}|$, $\tilde{\mathcal{D}}$ should contain similar disease types as $\mathcal{D}$.
\item \textit{Visit/patient-level Jensen-Shannon divergence (JSD$_{\{v, p\}}$)}: JSD is a metric to evaluate a visit/patient-level distribution of disease relative frequency between $\tilde{\mathcal{D}}$ and real EHR dataset $\mathcal{D}$. Here, the visit/patient-level frequency of a disease means the relative frequency of visit/patient that this disease appears. For patient-level frequency, if a disease appears in multiple visits of a patient, the disease frequency is still counted as 1. A lower divergence value means better generation quality.
\item \textit{Visit/patient-level normalized distance (ND$_{\{v, p\}}$)}: The Jensen-Shannon divergence focuses on the overall distributions, especially on the difference between data points that have high probability. As a result, the penalty should not be given for the difference between uncommon diseases that originally have a low probability. Therefore, to further evaluate the distribution of uncommon diseases, we adopt a normalized visit/patient-level distance. Given two distributions $p_{\boldsymbol{x}}$ and $p_{\tilde{\boldsymbol{x}}}$ of the visit/patient-level disease relative frequency in real and generated datasets, the distance is calculated as follows:
\begin{align}
    ND = \frac{1}{d}\sum_{i \in \mathcal{C}}{\frac{2|p_{\boldsymbol{x}}(i) - p_{\tilde{\boldsymbol{x}}}(i)|}{p_{\boldsymbol{x}}(i) + p_{\tilde{\boldsymbol{x}}}(i)}}.
\end{align}
Here, $\mathcal{C}$ is the entire disease set in the EHR dataset, and $d$ is the number of diseases as mentioned before. A good generation should also have a low normalized distance.
\item \textit{Required sample number to generate all diseases (RN)}: We use this metric to evaluate the ability of GANs to generate uncommon diseases. When generating all diseases, $|\tilde{\mathcal{D}}|$ should also contain close sample numbers to $|\mathcal{D}|$.
\end{itemize}

To assess whether the generated EHR dataset $\tilde{\mathcal{D}}$ is actually meaningful as an extension of real EHR data, we use a deep learning-based approach to train predictive models for health events on real and synthetic training data. More specifically, we pre-train predictive models on synthetic data, fine-tune these models on real training data, and finally test them on real test data of downstream tasks. It aims to quantify how much the generated EHR data can boost the training of predictive models on downstream tasks. Here, we apply three temporal prediction tasks:
\begin{itemize}[leftmargin=*]
    \item \textit{Diagnosis prediction}: It predicts all diagnoses of a patient in the visit $T + 1$ given previous $T$ visits. It is a multi-label classification.
    \item \textit{Heart failure/Parkinson's disease prediction}: It predicts if a patient will be diagnosed with heart failure/Parkinson's disease in the visit $T + 1$ given all the previous $T$ visits\footnote{The ICD-9 codes for heart failure and Parkinson's disease start with 428 and 332, respectively.}. It is a binary classification. Here, heart failure is one of the most frequent diseases in the EHR datasets we used in this work. The Parkinson's disease is an uncommon disease.
\end{itemize}

The evaluation metrics for diagnosis prediction are weighted F-1 score (w-$F_1$). For heart failure/Parkinson prediction, we use the area under the ROC curve (AUC).

\subsection{Datasets}
\begin{figure}
    \centering
    \subfloat[MIMIC-III]{\includegraphics[width=0.5\linewidth]{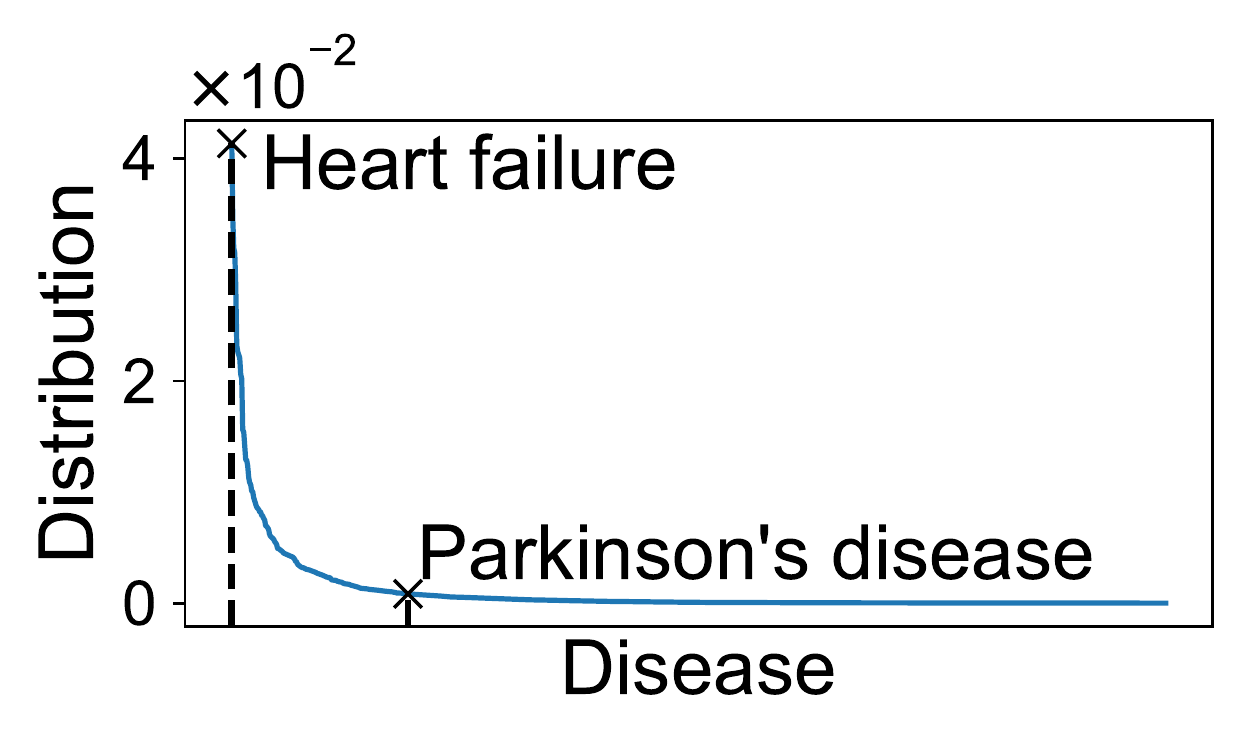}\label{fig:mimic3_stat}}
    \subfloat[MIMIC-IV]{\includegraphics[width=0.5\linewidth]{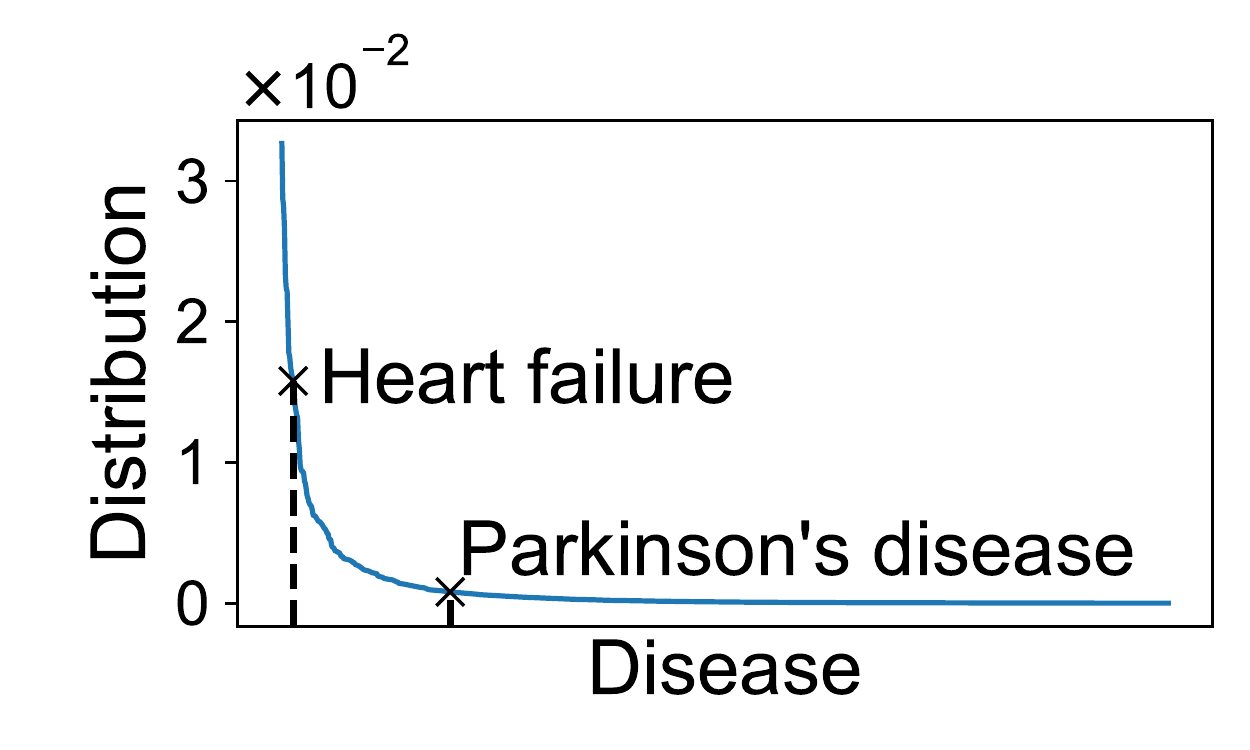}\label{fig:mimic4_stat}}
    \caption{Visit-level relative frequency distribution of diseases in the MIMIC-III and MIMIC-IV datasets.}
    \label{fig:mimic_stat}
\end{figure}

We use the MIMIC-III~\cite{johnson2016mimic} and MIMIC-IV~\cite{mimiciv} datasets to validate the generation of {\modelname}. MIMIC-III contains 7,493 patients who have multiple visits, i.e., visit sequences, from 2001 to 2012. For MIMIC-IV, we randomly select 10,000 patients with multiple visits from 2013 to 2019 to avoid overlaps with MIMIC-III. The statistics of MIMIC-III and MIMIC-IV are shown in Table~\ref{tab:dataset}. We also illustrate the visit-level disease distribution of MIMIC-III and MIMIC-IV in \figurename~\ref{fig:mimic_stat} in descending order, including annotations for heart failure and Parkinson's disease. The curves illustrate that the disease relative frequency in both datasets is a long-tail distribution. It further verifies the significance of improving the generation quality for uncommon diseases.

To conduct the predictive tasks, we randomly split the two datasets into training and test sets. The MIMIC-III contains 6,000 and 1,493 patients in training and test sets, respectively, while MIMIC-IV contains 8,000 and 2,000, respectively. The training sets of MIMIC-III and MIMIC-IV have 16,055 and 29,804 visits, respectively. It is worth noting that {\modelname} is trained using the training sets to ensure there is no data leakage when testing.

\begin{table}
    \centering
    \caption{Statistics of MIMIC-III and MIMIC-IV datasets.}
    \label{tab:dataset}
    \begin{tabular}{lcccc}
        \toprule
        \textbf{Dataset} & \textbf{MIMIC-III} & \textbf{MIMIC-IV} \\
        \midrule
        \# patients             & 7,493  & 10,000 \\
        \# visits               & 19,894 & 36,607 \\
        Max. \# visit per patient       & 42     & 55 \\
        Avg. \# visit per patient       & 2.66   & 3.66 \\
        \midrule
        \# diseases                & 4,880  & 6,102  \\
        Max. \# diseases per visit      & 39       & 50 \\
        Avg. \# diseases per visit      & 13.06       & 13.38 \\
        \midrule
        \# patients with heart failure & 3,364  & 2,137 \\
        \# patients with Parkinson & 109 & 153 \\
        \bottomrule
    \end{tabular}
\end{table}

\begin{table*}
    \centering
    \caption{Statistical evaluation results on generated data based on MIMIC-III and MIMIC-IV. GT: Generated disease type; JSD$_v$, JSD$_p$: Visit/patient-level Jensen-Shannon divergence; ND$_v$, ND$_p$: Visit/patient-level normalized distance; RN: Required sample number to generate all disease types. }
    \label{tab:stat_result}
    \begin{tabular}{clccccccccc}
        \toprule
        \multicolumn{2}{c}{\multirow{2}{*}{\textbf{Metrics}}} & \multicolumn{4}{c}{\textbf{Single Visit}} & \multicolumn{4}{c}{\textbf{Visit Sequence}} & \multirow{2}{*}{\textbf{Real}} \\ \cmidrule(lr){3-6}\cmidrule(lr){7-10}
        \multicolumn{2}{c}{~}
                     &    medGAN    &    CTGAN     &    EMR-WGAN  &   RDP-CGAN   &    WGAN-GP   &    TimeGAN   &    T-CGAN    &   {\modelname}    &        \\ 
        \midrule
        \multirow{7}{*}{\rotatebox[origin=c]{90}{MIMIC-III}}
        & GT         &    1,356     &    2,742     &    1,210     &    3,161     &    1,775     &    1,037     &    2,344     &   \textbf{4,431}  & 4,880  \\
        & JSD$_v$    &    0.2342    &    0.1983    &    0.1762    &    0.1587    &    0.1843    &    0.3344    &    0.1604    &   \textbf{0.1344} & 0      \\
        & JSD$_p$    &      ---     &      ---     &      ---     &      ---     &    0.2022    &    0.3518    &    0.1969    &   \textbf{0.1413} & 0      \\
        & ND$_v$     &    1.6751    &    1.2911    &    1.6213    &    0.9067    &    1.5817    &    1.7791    &    1.2943    &   \textbf{0.6563} & 0      \\
        & ND$_p$     &      ---     &      ---     &      ---     &      ---     &    1.5924    &    1.7719    &    1.3312    &   \textbf{0.6645} & 0      \\
        & RN         &    $>10^7$   &    $>10^7$   &    $>10^7$   &    $>10^7$   &    $>10^7$   &    $>10^7$   &    $>10^7$   &   \textbf{7,952}  & 6,000  \\
        & \# Params  &    3.84M     &    2.59M     &    1.96M     &    12.07M    &    1.35M     &    3.05M     &    1.95M     &   5.84M           & ---    \\
        \midrule
        \multirow{7}{*}{\rotatebox[origin=c]{90}{MIMIC-IV}}
        & GT         &    1,807     &    2,915     &    1,396     &    3,835     &    1,747     &    1,331     &    2,686     &   \textbf{5,677}  & 6,102  \\
        & JSD$_v$    &    0.2130    &    0.2217    &    0.1912    &    0.1662    &    0.2135    &    0.4004    &    0.1540    &   \textbf{0.1467} & 0      \\
        & JSD$_p$    &      ---     &      ---     &      ---     &      ---     &    0.2500    &    0.4153    &    0.1963    &   \textbf{0.1649} & 0      \\
        & ND$_v$     &    1.6709    &    1.4306    &    1.6902    &    0.9709    &    1.6911    &    1.7849    &    1.4222    &   \textbf{0.6705} & 0      \\
        & ND$_p$     &      ---     &      ---     &      ---     &      ---     &    1.7015    &    1.7911    &    1.4731    &   \textbf{0.6843} & 0      \\
        & RN         &    $>10^7$   &    $>10^7$   &    $>10^7$   &    $>10^7$   &    $>10^7$   &    $>10^7$   &    $>10^7$   &   \textbf{11,734} &  10,000 \\
        & \# Params  &    4.78M     &    3.21M     &    2.43M     &    15.01M    &    1.67M     &    3.68M     &    2.42M     &   7.25M          & ---    \\
        \bottomrule
    \end{tabular}
\end{table*}

\subsection{Baseline Models}
To evaluate the quality of generated EHR data, we select various GAN models as baselines. They can be divided into two major types: GANs for visit-level generation and GANs for patient-level generation.
\subsubsection{GANs for Visit-Level Generation}
We adopt four GANs as baselines that generate single visits:
\begin{itemize}[leftmargin=*]
    \item \textit{medGAN}~\cite{choi2017generating}: It uses the generator to output a latent feature and applies a pre-trained auto-encoder to decode the latent feature as the discriminator input.
    \item \textit{CTGAN}~\cite{xu2019modeling}: It uses a training-by-sampling strategy to generate imbalanced tabular data. The input of the CTGAN generator is the concatenation of a noise vector and a conditional vector.
    \item \textit{EMR-WGAN}~\cite{zhang2020ensuring}: It removes the auto-encoder in medWGAN~\cite{baowaly2019synthesizing} and directly generates visits with the generator.
    \item \textit{RDP-CGAN}~\cite{torfi2022differentially}: It uses a convolutional auto-encoder and discriminator under the framework of WGAN.
\end{itemize}
For these GANs generating single visits, we only calculate GT, JSD$_v$, ND$_v$, and RN to evaluate the statistical results and do not use them for temporal prediction tasks since they cannot generate visit sequences.

\subsubsection{GANs for Patient-Level Generation}
We select three GANs to generate visit sequences:
\begin{itemize}[leftmargin=*]
    \item \textit{WGAN-GP}~\cite{gulrajani2017improved}: We implement WGAN-GP's generator with GRU to generate time-series data. The input of each GRU cell is a random noise.
    \item \textit{TimeGAN}~\cite{yoon2019time}: It uses RNN to generate hidden features and proposes unsupervised, supervised, and reconstruction losses to train the GAN model.
    \item \textit{T-CGAN}~\cite{ramponi2018t}: It applies a conditional GAN by specifying the time step of generated visits. The input of the generator is a concatenation of a noise vector and a time step conditional vector. When generating visit sequences, we specify the time step from 1 to $T$ and combine all generated visits chronologically into a sequence.
\end{itemize}

\subsection{Parameter Settings}
The parameter settings for baselines are listed as follows:
\begin{itemize}[leftmargin=*]
    \item medGAN: We use three fully-connected (FC) layers with skip-connection and batch normalization as the generator. Each layer has 128 hidden units. The discriminator has three FC layers with 256 and 128 hidden units. The autoencoder contains two FC layers with 128 hidden units.
    \item CTGAN: It uses three FC layers without skip-connection as the generator. The hidden units are all 128. The discriminator is the same as medGAN.
    \item EMR-WGAN: The generator and critic of EMR-WGAN have the same hyper-parameter settings as medGAN.
    \item RDP-CGAN: We use six 1-d conv layers for both encoder and decoder with kernel sizes \{3, 3, 4, 4, 4, 4\} and \{4, 4, 4, 4, 3, 3\}. We use three conv layers in the generator with kernel sizes \{3, 3, 3\} and five conv layers in the critic with kernel sizes \{3, 3, 4, 4, 4\}.
    \item WGAN-GP: It uses a GRU with 128 hidden units as the generator. The critic has two FC layers with 128 hidden units. We calculate the sum of the Wasserstein distance for each visit as the final discriminator loss.
    \item TimeGAN: The generator, discriminator, embedder, recovery, and supervisor all have a GRU with 128 hidden units.
    \item T-CGAN: It has the same generator and critic as CTGAN.
\end{itemize}
In our experiments, we ran {\modelname} multiple times and investigated the model performance with different randomly initialized parameters. We found that the model tends to provide results at the same level under different random initializations. Therefore, we randomly initialize all model parameters to achieve generality. The size of the noise vector as well as GRU hidden units $s$ is 256. The MLP used in the critic has one hidden layer with 64 hidden units. For base GRU pre-training, we run 200 epochs with Adam optimizer~\cite{KingmaB14} and set the learning rate to $10^{-3}$. For training {\modelname}, we run $3\times10^5$ iterations with batch size 256. The learning rates for the generator and critic are $10^{-4}$ and $10^{-5}$ and decay by 0.1 every $10^{5}$ iterations. The critic training number ${n}_{\text{critic}}$ is 1. We use the Adam optimizer and set $\beta_1 = 0.5$ and $\beta_2 = 0.9$. The $\lambda$ for gradient penalty is 10, the same as WGAN-GP~\cite{gulrajani2017improved}. All programs are implemented using Python 3.8.6 and PyTorch 1.9.1 with CUDA 11.1 on a machine with Intel i9-9900K CPU, 64GB memory, and Geforce RTX 2080 Ti GPU. The source code of {\modelname} is released publicly at \url{https://github.com/LuChang-CS/MTGAN}.

%% file: sections/experiment.tex
\section{Experimental Results}
\label{sec:exp}
\subsection{Statistical Evaluation}
To evaluate the statistical difference between generated EHR data $\tilde{\mathcal{D}}$ and real data $\mathcal{D}$, we utilize visit-level GANs to generate 16,055 and 29,084 visits, and utilize patient-level GANs to generate 6,000 and 10,000 patients, when training with MIMIC-III and MIMIC-IV, respectively. The statistical evaluation results on these datasets are shown in Table~\ref{tab:stat_result}.

For the generated disease types (GT), the results should be close to real disease types. All baselines can only generate less than 4,000 diseases, while the disease types generated by {\modelname} are close to real data. The visit/patient-level Jensen-Shannon divergence (JSD$_v$, JSD$_p$) shows that {\modelname} can synthesize a good EHR dataset in terms of the overall disease distribution, while the results of other baselines are almost on par. However, when considering uncommon diseases, we can conclude from the normalized distance (ND$_v$, ND$_p$) that {\modelname} has better ability in generating diseases with low frequency than other baselines. This conclusion is further validated by the required sample number (RN) to generate all diseases. In this experiment, we keep generating samples until the disease type in the synthetic dataset $\tilde{\mathcal{D}}$ reaches the disease type in the real dataset $\mathcal{D}$. For all baselines, we stop at $10^7$ samples given that they cannot generate more uncommon diseases. However, {\modelname} is able to generate all diseases only using 7,952 and 11,734 samples for MIMIC-III and MIMIC-IV, respectively. Although these sample numbers are larger than the real patient numbers in MIMIC-III and MIMIC-IV, the ability to generate uncommon diseases of {\modelname} is verified.

When comparing visit-level and patient-level distance of GANs for visit sequences, it should be noted that almost all models have lower scores for JSD$_v$ than JSD$_p$. It shows that retaining temporal correlation in visit sequences is harder than solely learning the disease distribution in single visits. In spite of this, we see that {\modelname} has a minimal difference between JSD$_v$ and JSD$_p$ than nearly all baselines. It is worth noting that although T-CGAN also achieves a relative low distance, it has a large difference between visit-level and patient-level distance. We infer that T-CGAN does not keep temporal information in visit sequences, because it only specifies the time step of a visit while not considering previous visits. Therefore, we can conclude that {\modelname} is able to generate visit sequences, and meanwhile preserve temporal correlations between adjacent visits.

In summary, {\modelname} can generate uncommon diseases as well as keep similar disease distribution to real EHR datasets. Meanwhile, {\modelname} can generate visit sequences while considering temporal correlations between visits.

\subsection{Analysis of GAN Training}

\begin{figure}
    \centering
    \subfloat[MIMIC-III]{\includegraphics[width=\linewidth]{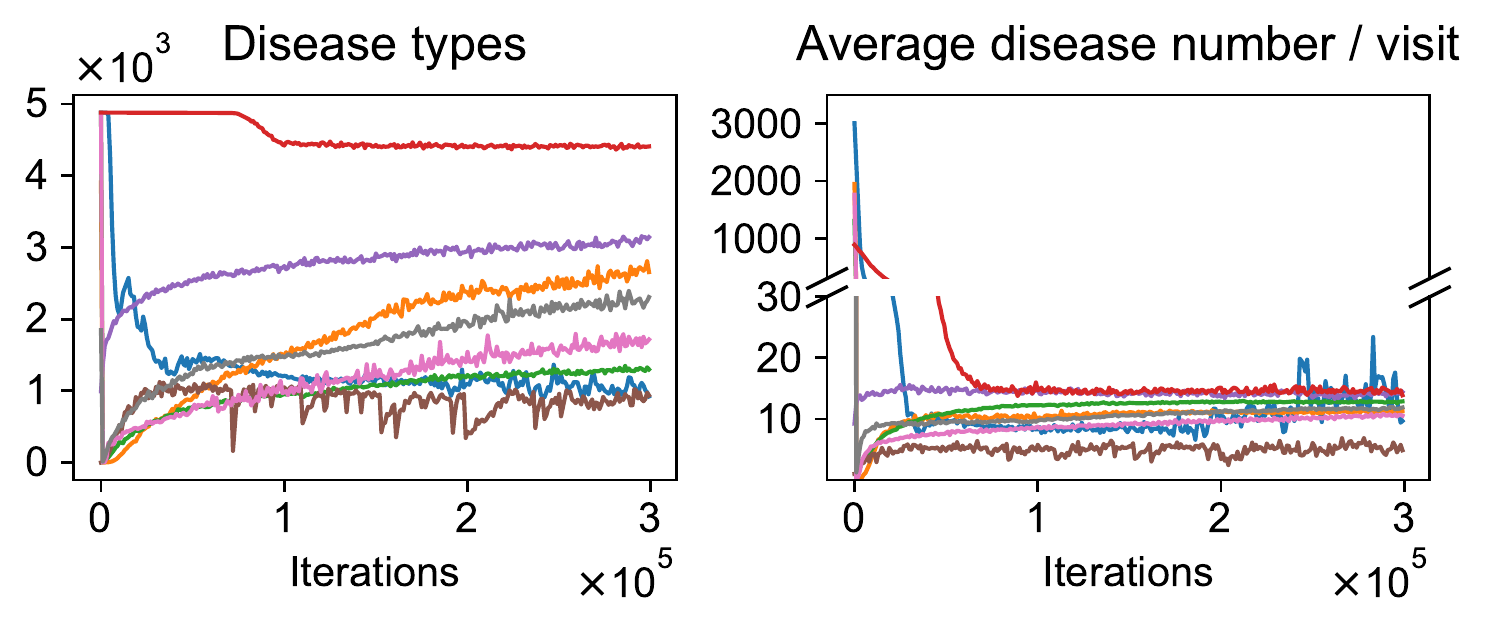}\label{fig:history_mimic3}} \\
    \subfloat[MIMIC-IV]{\includegraphics[width=\linewidth]{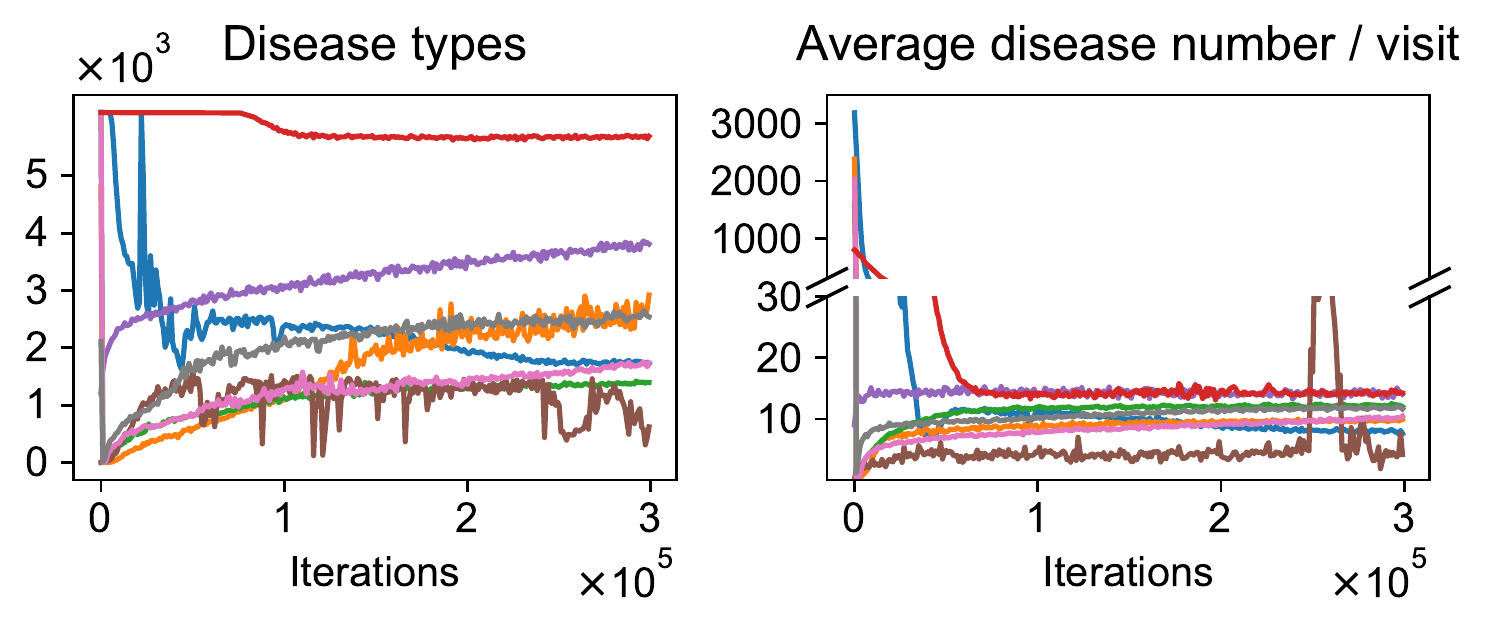}\label{fig:history_mimic4}} \\
    \subfloat{\includegraphics[width=\linewidth]{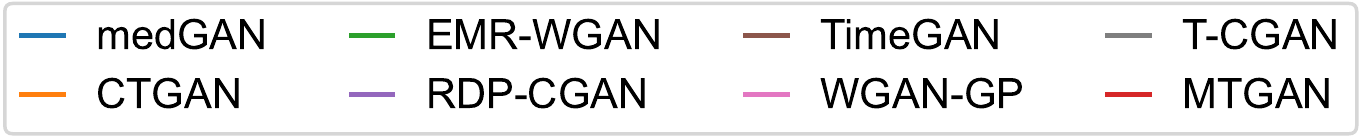}}
    \caption{The trend of generated disease types (left) and average disease number per visit (right) during training on MIMIC-III and MIMIC-IV.}
    \label{fig:history}
\end{figure}

To analyze the stability of training different GANs for generating EHR data, we plot the disease types and average disease number per visit during training on the MIMIC-III and MIMIC-IV datasets in \figurename~\ref{fig:history}. From the trend of disease types in the left figures of \figurename~\ref{fig:history}\subref{fig:history_mimic3} and \figurename~\ref{fig:history}\subref{fig:history_mimic4}, we notice that baseline GANs relying on the Wasserstein distance (CTGAN, EMR-WGAN, WGAN-GP, and T-CGAN) can generate over 2,000 disease types at the beginning of training. However, this number dramatically drops to 0 after a few iterations, and slowly increases during training. This phenomenon can also be reflected by the average disease number per visit in the right figures. The average disease number per visit of these GANs starts from a high value (over 1,000). Then this number decreases to 0 and increases around the real average disease number of MIMIC-III and MIMIC-IV, i.e., 13.06 and 13.38. We infer it is due to the instability of baseline GANs. In the beginning of training, the generator of baselines tend to generate zero diseases in order to get a lower Wasserstein distance, because of the high sparsity of the EHR data. An exception is RDP-CGAN. We infer it is because RDP-CGAN pre-trains the auto-encoder so that it can generate diseases in the beginning. On the contrary, GANs based on the binary classification (medGAN, TimeGAN) do not have such a phenomenon but they are unstable at a latter stage. Nevertheless, they converge at a lower number of disease types and have a higher Jensen-Shannon divergence and normalized distance.

However, from \figurename~\ref{fig:history}, we can see that the generated disease types of {\modelname} stabilize at a high number and are close to the real number of disease types in MIMIC-III (4,880) and MIMIC-IV (6,102). Furthermore, even though {\modelname} is also based on the Wasserstein distance, the average disease number per visit does not dramatically drop, but gradually decreases to the real data. This makes {\modelname} more stable and easier to train than baselines, in terms of adjusting the learning rate, batch size, and other key hyper-parameters.

\begin{table}
    \centering
    \caption{Statistical results of {\modelname} variants on MIMIC-III and MIMIC-IV.}
    \label{tab:ablation}
    \begin{tabular}{clccccc}
        \toprule
        & \textbf{Metrics} & {\modelshortname}$_\text{h-}$ & {\modelshortname}$_\text{c-}$ & {\modelshortname}$_\text{dist}$ & {\modelshortname}$_\text{trans}$ & {\modelname} \\
        \midrule
        \multirow{6}{*}{\rotatebox[origin=c]{90}{MIMIC-III}}
        & GT      & 4,284 & 4,044 & 3,339 & 3,362 & \textbf{4,431} \\
        & JSD$_v$ & 0.4167 & 0.1508 & 0.1414 & 0.1791 & \textbf{0.1344} \\ 
        & JSD$_p$ & 0.4040 & 0.1534 & 0.1521 & 0.1846 & \textbf{0.1413} \\
        & ND$_v$  & 0.7637 & 0.8467 & 0.9894 & 1.0244 & \textbf{0.6563} \\
        & ND$_p$  & 0.7783 & 0.8388 & 0.9959 & 1.0038 & \textbf{0.6645} \\
        & RN      & 9,232  & $>10^7$ & 229,628 & 170,800 & \textbf{7,952} \\
        \midrule
        \multirow{6}{*}{\rotatebox[origin=c]{90}{MIMIC-IV}}
        & GT      & 5,622   & 4,548   & 4,588   & 4,330   & \textbf{5,677}  \\
        & JSD$_v$ & 0.3327  & 0.1910  & 0.1769  & 0.1599  & \textbf{0.1467}  \\ 
        & JSD$_p$ & 0.3264  & 0.2028  & 0.1935  & 0.1957  & \textbf{0.1649}  \\
        & ND$_v$  & 0.7711  & 0.9748  & 0.9460  & 1.0586  & \textbf{0.6705}  \\
        & ND$_p$  & 0.7961  & 0.9962  & 0.9783  & 1.0619  & \textbf{0.6843}  \\
        & RN      & 40,850  & $>10^7$ & 1,488,966  & 91,530  & \textbf{11,734}  \\
        \bottomrule
    \end{tabular}
\end{table}

\begin{table}
    \centering
    \caption{Time taken (in seconds) for training one iteration and generating 6,000 samples.}
    \label{tab:time}
    \begin{tabular}{ccccc}
    \toprule
    Model & WGAN-GP & TimeGAN & T-CGAN & \modelname \\
    \midrule
    Training & 0.15 & 0.41 & 0.19 & 0.23 \\
    Generating & 2.36 & 5.87 & 2.57 & 3.02 \\
    \bottomrule
    \end{tabular}
\end{table}

\begin{table*}
\centering
\caption{
Downstream task evaluation by pre-training Dipole and GRAM on synthetic data, fine-tuning on real training data. The results are reported on real test data. Note that ``w/o synthetic'' indicates that the model is trained using only real training data. We use w-$F_1$ (\%) for Diagnosis prediction, AUC (\%) for Heart failure and Parkinson's disease prediction. The synthetic data have equal sample numbers as real training data.}
\label{tab:downstream}
\begin{tabular}{clcccccc}
    \toprule
    & \multirow{2}{*}{\textbf{Models}} & \multicolumn{3}{c}{\textbf{Dipole}} & \multicolumn{3}{c}{\textbf{GRAM}} \\ \cmidrule(lr){3-5}\cmidrule(lr){6-8}
    \multicolumn{2}{c}{~}
    &    Diagnosis    &    Heart Failure     &    Parkinson   &   Diagnosis     &    Heart Failure      &    Parkinson  \\ 
    \midrule
    \multirow{5}{*}{\rotatebox[origin=c]{90}{MIMIC-III}} &
    w/o synthetic & 19.35 & 82.08 & 68.80 & 21.52 & 83.55 & 73.81 \\
    & w/ WGAN-GP & 20.02 (+3.46\%) & 82.67 (+0.72\%) & 69.11 (+0.45\%) & 22.48 (+4.46\%) & 84.06 (+0.61\%) & 74.29 (+0.65\%) \\
    & w/ TimeGAN & 19.60 (+1.29\%) & 82.69 (+0.74\%) & 68.57 (-0.33\%) & 22.06 (+2.51\%) & 83.84 (+0.35\%) & 73.85 (+0.05\%) \\
    & w/ T-CGAN & 20.38 (+5.32\%) & 83.38 (+1.58\%) & 69.33 (+0.77\%) & 22.30 (+3.62\%) & \textbf{84.22 (+0.80\%)} & 74.40 (+0.80\%) \\
    & w/ {\modelname} & \textbf{20.48 (+5.84\%)} & \textbf{83.41 (+1.62\%)} & \textbf{70.45 (+2.40\%)} & \textbf{22.57 (+4.88\%)} & 84.19 (+0.77\%) & \textbf{75.06 (+1.69\%)} \\
    \midrule
    \multirow{5}{*}{\rotatebox[origin=c]{90}{MIMIC-IV}} &
    w/o synthetic       & 23.69  & 88.69  & 72.59  & 23.50  & 89.61  & 78.51  \\
    & w/ WGAN-GP      & 24.17 (+2.03\%)  & 88.78 (+0.10\%)  & 72.81 (+0.30\%)  & 23.68 (+0.77\%)  & 89.81 (+0.22\%)  & 78.81 (+0.38\%)  \\
    & w/ TimeGAN      & 23.62 (-0.30\%)  & 88.63 (-0.07\%)  & 72.55 (-0.06\%)  & 23.61 (+0.47\%)  & 89.68 (+0.08\%)  & 78.56 (+0.06\%) \\
    & w/ T-CGAN       & 24.60 (+3.84\%)  & 89.04 (+0.39\%)  & 72.76 (+0.23\%)  & 23.75 (+1.06\%)  & 89.94 (+0.37\%)  & 78.90 (+0.50\%)  \\
    & w/ {\modelname} & \textbf{24.74 (+4.43\%)}  & \textbf{89.11 (+0.47\%)}  & \textbf{73.16 (+0.79\%)}  & \textbf{24.09 (+2.51\%)}  & \textbf{90.05 (+0.49\%)}  & \textbf{79.35 (+1.07\%)} \\
    \bottomrule
\end{tabular}
\end{table*}

\subsection{Empirical Time Complexity Analysis}

To further demonstrate the time required in training and generating, we report the time in seconds (s) of the GANs for visit sequence generation in Table~\ref{tab:time}, i.e., WGAN-GP, TimeGAN, T-CGAN, and the proposed \modelname. Here, the training time refers to the time for training models in an iteration, while generating time is the time for generating 6,000 samples. We see that although with the conditional matrix and sampling strategies, both the training and generating time of {\modelname} are at the same level compared with other methods, which means that our model can achieve better performance without increasing the time complexity.

\subsection{Ablation Study}
To study the effectiveness of the various components, we conduct ablation studies by removing or changing parts of the model. The variants of {\modelname} are listed as follows:
\begin{itemize}[leftmargin=*]
    \item {\modelshortname}$_\text{h-}$: In the critic, we remove the hidden state in Equation~(\ref{eq:critic_cat}). In addition, we let the generator only output the probability but not the hidden state of GRU.
    \item {\modelshortname}$_\text{c-}$: We remove the conditional matrix in the generator to verify the contribution of it to uncommon disease generation. As a result, the generated synthetic data are directly sampled from the GRU outputs.
    \item {\modelshortname}$_{\text{dist}}$: In Equations~(\ref{eq:loss_d}) and (\ref{eq:loss_g}), we uniformly sample target diseases. In {\modelshortname}$_{\text{dist}}$, we sample target diseases from the visit-level disease distribution in real EHR dataset to study the impact of sampling in the GAN training.
    \item {\modelshortname}$_{\text{trans}}$: To test the effect of GRU in the generator of {\modelname}, we replace $g_{\text{gru}}$ with Transformer~\cite{vaswani2017attention}, since Transformer is also effective in EHR-related tasks~\cite{meng2021bidirectional,amin2020exploring,li2020behrt}. More specifically, we use a Transformer encoder module, including a positional encoding part and a masked self-attention part to generate diseases from $T$ noises. In the critic, we also remove the hidden state in Equation~(\ref{eq:critic_cat}), since the generator cannot output it for synthetic data.
\end{itemize}

We report the statistical results of {\modelname} variants in Table~\ref{tab:ablation}. Comparing {\modelshortname}$_\text{h-}$ and {\modelname}, we notice both JSD and ND have a large increase, but it can still generate all disease types within a small sample number. However, after removing the conditional matrix, {\modelshortname}$_\text{c-}$ cannot generate all disease types with $10^7$ generated samples. We can conclude that distinguishing hidden states in the critic is able to improve the quality of synthetic EHR data in terms of the disease distribution, and the conditional matrix helps to learn the distribution of uncommon diseases.

When comparing between {\modelshortname}$_{\text{dist}}$ and {\modelname}, we notice that JSD does not have a large difference, but ND of {\modelshortname}$_{\text{dist}}$ increases a lot. Additionally, {\modelshortname}$_{\text{dist}}$ requires more samples to generate all disease types. We conjecture it is because uncommon diseases have low frequencies and therefore occur less in the synthetic data when sampling from the visit-level disease distribution. This also leads to a high normalized distance and more samples to generate all disease types.

The last comparison is replacing GRU in the generator with a Transformer encoder. Although Transformer is effective and has gained great success in natural language processing, it does not achieve superior performance to GRU. We infer that it is because the visit sequences in MIMIC-III and MIMIC-IV are not sufficiently long and hence GRU can adequately capture the temporal features of EHR data. Furthermore, we think even with positional encoding, it is still hard to learn temporal information given that the inputs of all time steps are noises.

In summary, we conclude that both the hidden state critique and the conditional matrix contribute to the EHR data generation in terms of overall disease distributions and especially uncommon diseases.

\begin{figure*}
    \centering
    \subfloat[Dipole on MIMIC-III]{\includegraphics[width=0.49\linewidth]{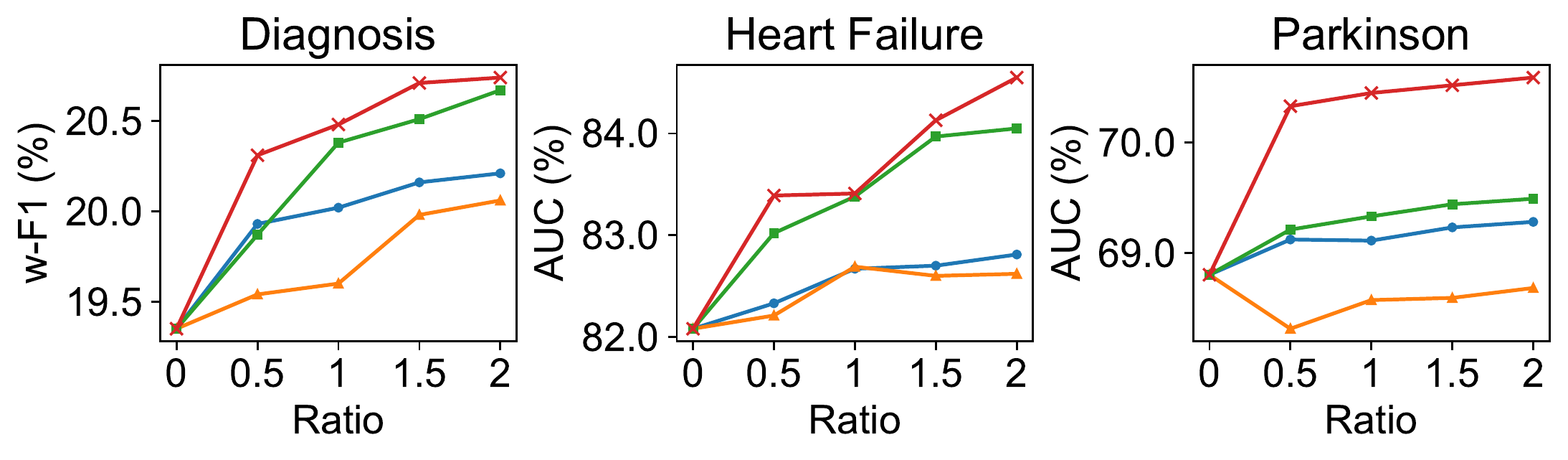}\label{fig:dipole_mimic3}} \hfill
    \subfloat[GRAM on MIMIC-III]{\includegraphics[width=0.49\linewidth]{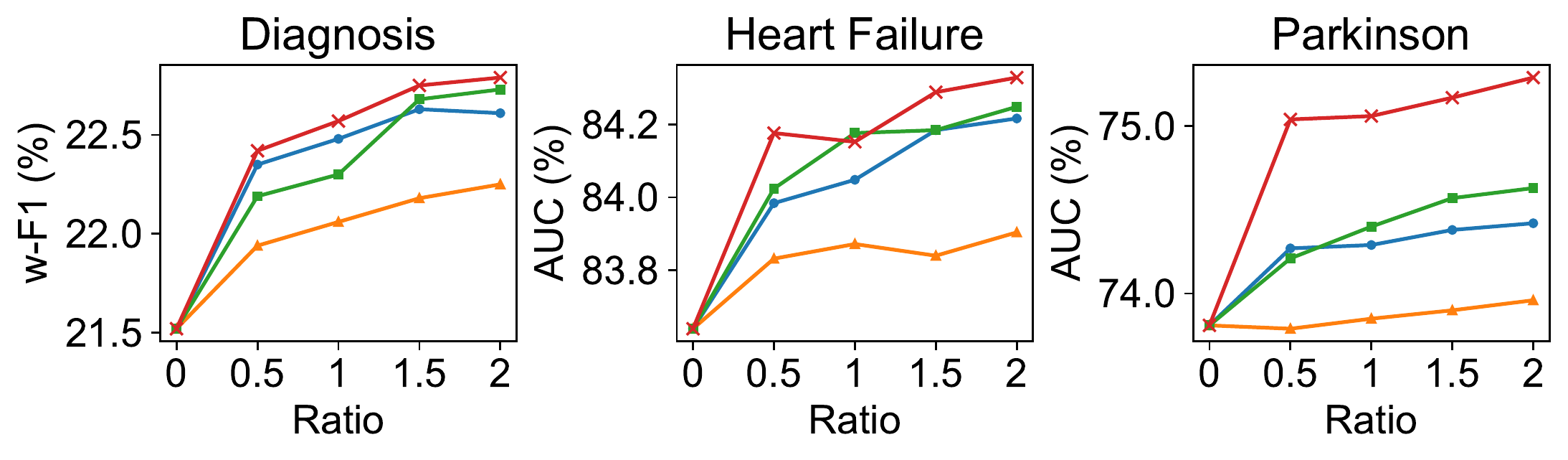}\label{fig:gram_mimic3}} \\
    \subfloat[Dipole on MIMIC-IV]{\includegraphics[width=0.49\linewidth]{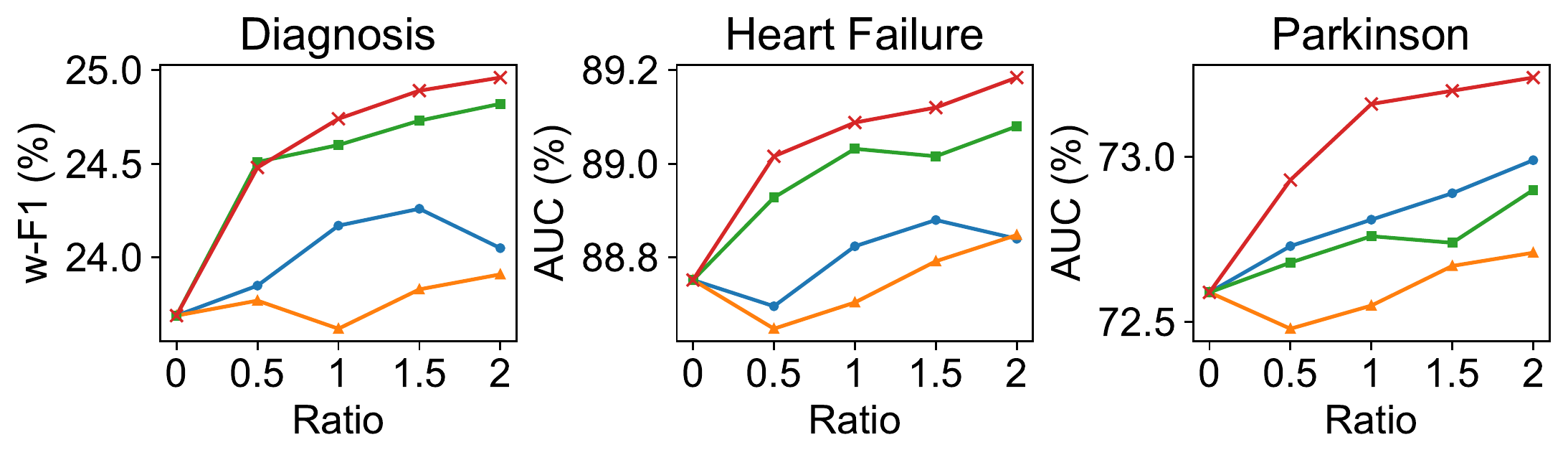}\label{fig:dipole_mimic4}} \hfill
    \subfloat[GRAM on MIMIC-IV]{\includegraphics[width=0.49\linewidth]{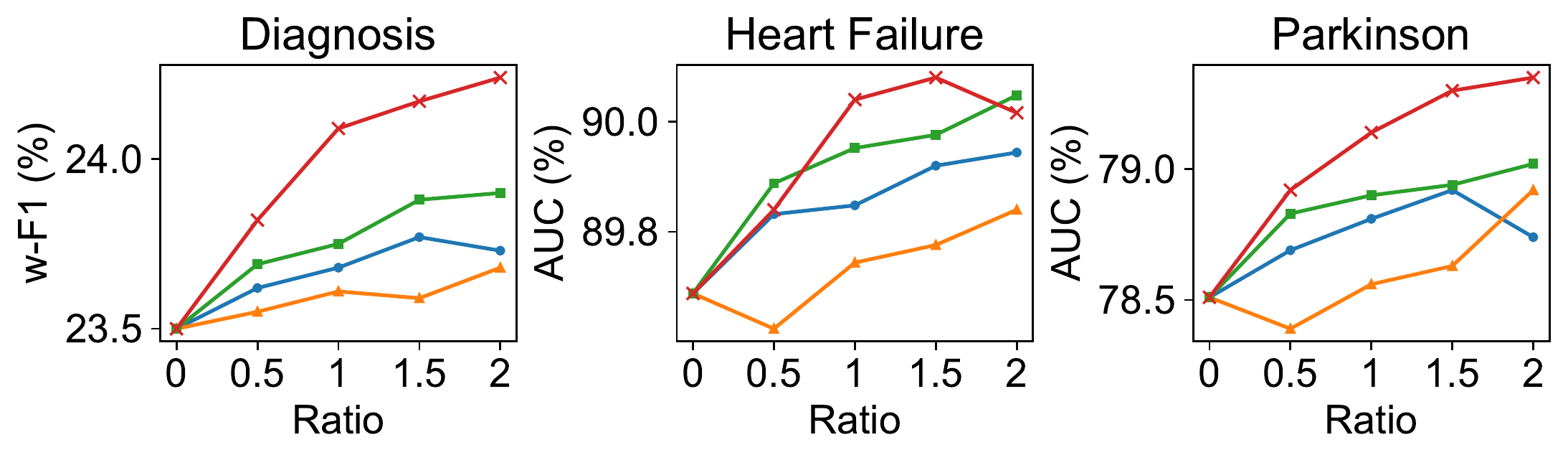}\label{fig:gram_mimic4}} \\
    \subfloat{\includegraphics[width=0.5\linewidth]{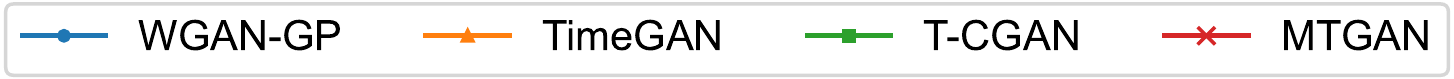}}
    \caption{Downstream task evaluation by training Dipole and GRAM with different ratios of synthetic data over real training data.}
    \label{fig:downstream}
\end{figure*}

\subsection{Downstream Task Evaluation}

In this experiment, we evaluate the synthetic data of GANs for patient-level generation, i.e., WGAN-GP, TimeGAN, T-CGAN, and {\modelname}. As mentioned in Section~\ref{sec:metrics}, we select three temporal prediction tasks as the downstream tasks: Diagnosis prediction, heart failure prediction, and Parkinson's disease prediction. Here, we choose two predictive models as baselines of downstream tasks:
\begin{itemize}[leftmargin=*]
    \item Dipole~\cite{ma2017dipole}: It is a bi-directional RNN with attention methods to predict diagnoses.
    \item GRAM~\cite{choi2017gram}: It is an RNN-based model using disease domain knowledge to predict diagnoses and heart failure.
\end{itemize}

We first train Dipole and GRAM only using the training data of MIMIC-III and MIMIC-IV as baselines (w/o synthetic). Then, we generate synthetic EHR that are trained with WGAN-GP, TimeGAN, T-CGAN, and {\modelname}, respectively. Here, the synthetic data have equal sample numbers to real training data. Next, we pre-train a new Dipole and GRAM using these synthetic data, fine-tune them using real training data, and finally test them on real test data. The experimental results including baseline results, pre-training and fine-tuning results, and their increments are shown in Table~\ref{tab:downstream}. In this table, the synthetic data can enhance the predictive models on almost all tasks, among which {\modelname} has the largest improvement on the diagnosis prediction. We infer that the synthetic EHR data provide more samples especially samples with uncommon diseases, so that the predictive models can provide a better prediction for them. Additionally, compared to other GANs, {\modelname} has the most predominant results on the Parkinson's disease prediction. It further proves that {\modelname} can learn a better distribution for uncommon diseases and boost downstream tasks especially related to these uncommon diseases.

In Table~\ref{tab:downstream}, we pre-train Dipole and GRAM on the synthetic data that have the same number of samples as that of the real training data. In addition, we conduct more experiments by adopting different sample numbers of synthetic data in pre-training. Here, we set the ratios of synthetic data over real training data as $\{0.5, 1, 1.5, 2\}$ to explore the impact of the synthetic data amount during pre-training on downstream tasks. The results are illustrated in \figurename~\ref{fig:downstream}. Each predictive model is tested on three tasks with every dataset. In general, with the growth of pre-training data, we notice that the w-$F_1$ and AUC of prediction also show an increasing trend. It shows that these GAN models can learn effective disease distributions in EHR data to some extent. It is still worth noting that {\modelname} can generate synthetic EHR data that are more beneficial to downstream tasks than other GANs, especially the Parkinson's disease prediction. Therefore, we may conclude that  {\modelname} can generate EHR data that have more accurate disease distribution and more advantages in boosting downstream tasks.